\documentclass[letterpaper]{article} 
\usepackage[preprint]{aaai2027}  
\usepackage[hyphens]{url}  
\usepackage{graphicx} 
\usepackage{natbib}  
\usepackage{caption} 
\usepackage{algorithm}
\usepackage{algorithmic}
\usepackage{booktabs}
\usepackage{amssymb} 
\usepackage{multirow}
\usepackage{array}

\title{PerFact: Perception-Derived Fact Prompting for 3D Brain MRI Report Generation}

\author{
Jianyu Sun\textsuperscript{\rm 1,*}\quad
Zhenxuan Zhang\textsuperscript{\rm 1,2,*}\quad
Guang Yang\textsuperscript{\rm 1,2,3,$\dagger$}\quad
Peter J. Lally\textsuperscript{\rm 1,$\dagger$}
}

\affiliations{
\textsuperscript{\rm 1}Department of Bioengineering, Imperial College London, London, UK\\
\textsuperscript{\rm 2}Imperial-X, Imperial College London, London, UK\\
\textsuperscript{\rm 3}School of Biomedical Engineering \& Imaging Sciences, King's College London, London, UK\\[1pt]
\textsuperscript{*}These authors contributed equally to this work. \quad
\textsuperscript{$\dagger$} Corresponding authors.
}

\begin{document}

\maketitle

\begin{abstract}
Radiology report generation has matured almost entirely on 2D chest radiographs, where the default route to better reports is a larger backbone or a pre-training one on medical data. We revisit that assumption on 3D multi-sequence brain MRI, a volumetric multi-disease regime, and find that the model is not the lever. Zero-shot medical and radiology vision-language models transfer poorly to brain MRI, with chest radiograph specialists failing most conspicuously, and five backbones fine-tuned identically across three model families and an order of magnitude in scale differ only marginally. What determines the quality of the report is the information injected into the prompt. We delegate perception to upstream 3D segmentation and classification, serialize their outputs into a structured fact sentence, and prompt a LoRA-adapted vision-language model with it; we call this \textbf{PerFact}. In a controlled study that fixes the backbone, data split, target reports, and adaptation while varying only the injected grounding, perception-derived facts outperform retrieved prior reports, retrieval becomes redundant once facts are present, and end-to-end predicted facts remain effective without any ground-truth annotation at inference. The residual gap between predicted and oracle facts is explained by the granularity of the facts rather than by the generator. Closed-ended visual question answering comes at no measurable cost to report quality, though the grounding source has little effect on it. On 3D brain MRI, grounding information, not model choice, is the dominant controllable factor in report quality.

\end{abstract}


\section{Introduction}
\begin{figure}[!t]
  \centering
  {\small\textbf{(a) Three of 496 slices reach the model}}\\[2pt]
  \includegraphics[width=\columnwidth]{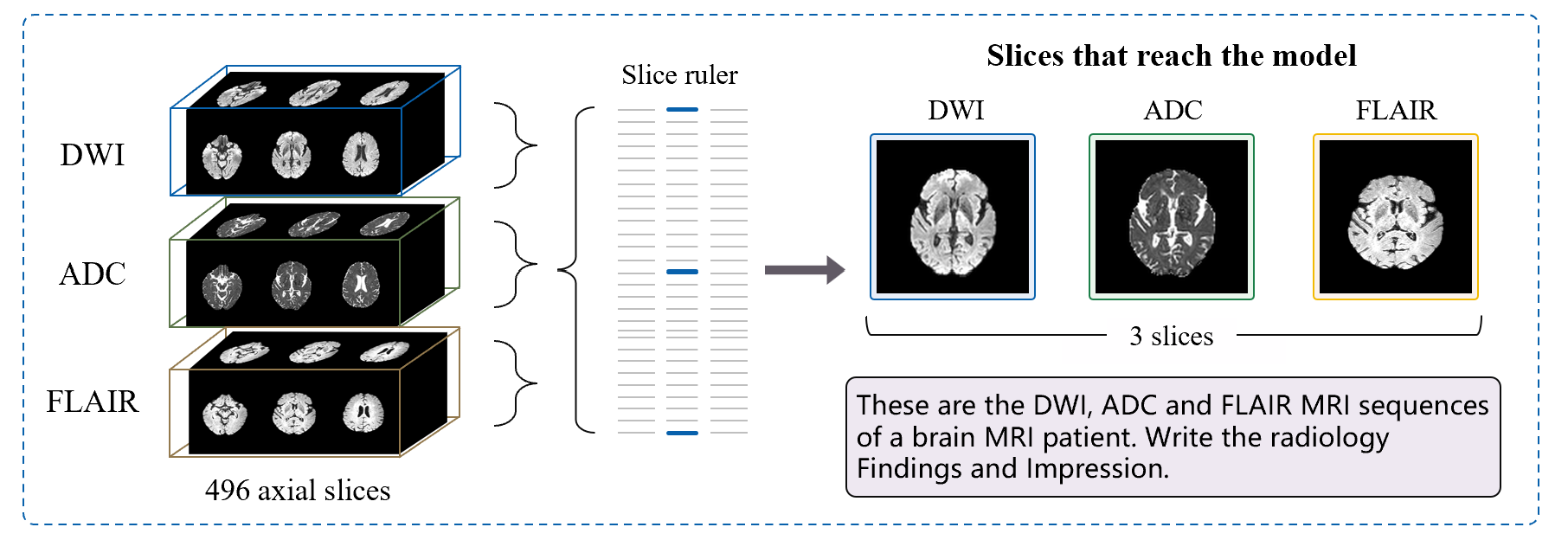}\\[7pt]
  {\small\textbf{(b) PerFact adds one machine-generated fact sentence}}\\[2pt]
  \includegraphics[width=\columnwidth]{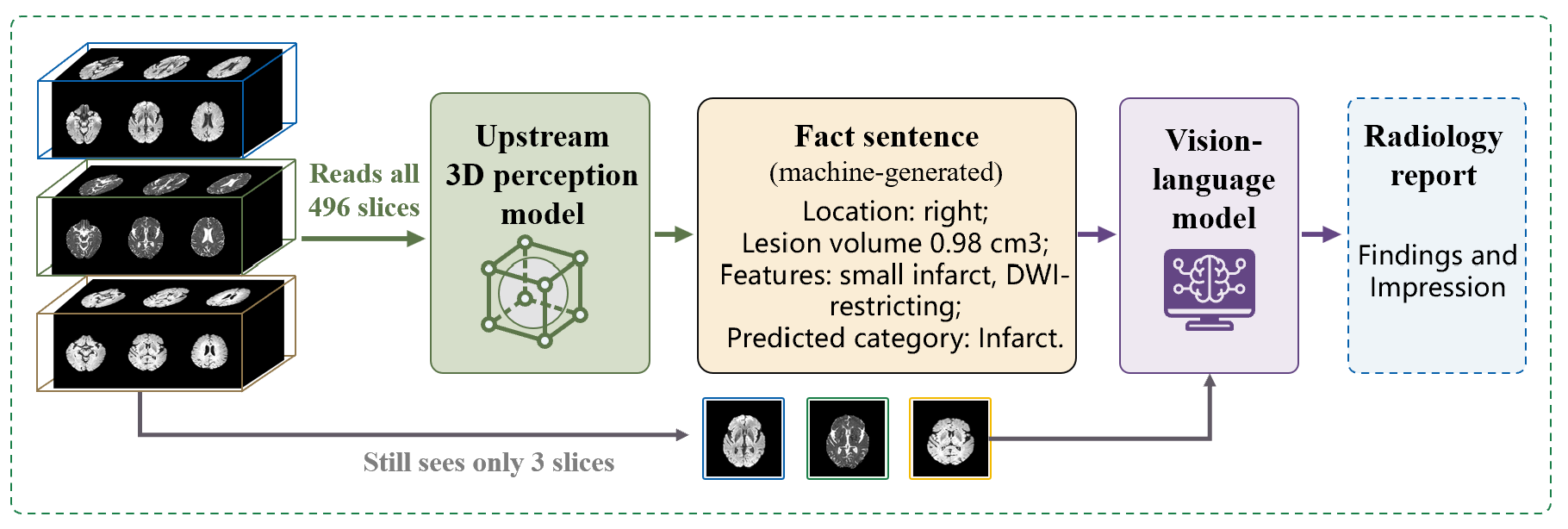}
  \caption{\textbf{The information gap PerFact closes.} This acute stroke study
  holds 496 axial slices and the prompt carries three of them, 0.6\% of the study,
  with a fixed instruction. The fact sentence comes from an upstream model that
  reads all 496; the vision-language model still sees only the same three slices.}
  \label{fig:teaser}
\end{figure}

A brain Magnetic Resonance Imaging (MRI) report is read for a few findings that change
management: where a lesion is, which side it is on, whether structures are displaced and in
which direction, and what the lesion most likely is. A vision-language model is asked for
those findings from three slices of a study that holds several hundred
(Figure~\ref{fig:teaser}). Fluent text that is wrong on any of
these carries no mark of its own unreliability. Figure~\ref{fig:hallucination} shows released
systems describing the wrong organ, placing a right frontal glioblastoma in the left
hemisphere, and inverting the direction of midline shift, all in well-formed prose.

Automated radiology report generation has matured on a single modality: the systems that define the state of the art were built, trained and validated on 2D chest radiographs \citep{maira2,llavarad,chexagent}. Brain MRI is a different regime. A study comprises several co-registered 3D volumes with different tissue contrasts rather than one projection, findings are volumetric and lateralized, and a realistic corpus spans diseases as different as glioma, meningioma, acute stroke and white-matter hyperintensity. Recent work shows that a health-system-scale neuroimaging archive can support strong report generation \citep{neurovfm}; what drives report quality in this regime, at the corpus sizes available to most groups, has not been established.

\definecolor{gtgreen}{HTML}{009E73}
\definecolor{halred}{HTML}{D55E00}
\definecolor{factblue}{HTML}{0072B2}
\definecolor{pathpurple}{HTML}{CC79A7}
\definecolor{factbg}{HTML}{EAF1F8}
\newcommand{\okmark}{\textcolor{gtgreen}{\textbf{\checkmark}}}
\newcommand{\badmark}{\textcolor{halred}{\textbf{$\times$}}}
\newcommand{\gt}[1]{\textcolor{gtgreen}{#1}}
\newcommand{\hl}[1]{\textcolor{halred}{#1}}
\newcommand{\el}{\textcolor{black!42}{\,\dots\,}}
\newsavebox{\hctab}
\newsavebox{\hcscan}
\newsavebox{\hcfact}
\newsavebox{\hcleft}
\newlength{\hctabh}
\newlength{\hcslack}
\newlength{\hcsetting}
\newlength{\hcleftw}
\newlength{\hcrightw}

\begin{figure*}[t]
  \centering
  \scriptsize
  \setlength{\hcleftw}{0.26\linewidth}%
  \setlength{\hcrightw}{0.71\linewidth}%
  \settowidth{\hcsetting}{Qwen2.5-VL (zero-shot)~\badmark}%
  \sbox{\hctab}{%
      \setlength{\tabcolsep}{0pt}%
      \renewcommand{\arraystretch}{1.527}%
      \begin{tabular}[t]{@{}l@{\hspace{7pt}}>{\centering\arraybackslash}p{\dimexpr\hcrightw-\hcsetting-7pt\relax}@{}}
        \toprule
        \textbf{Setting} & \textbf{Verbatim output on this study} \\
        \midrule
        \textbf{Reference}
          & Abnormal signal in the \gt{right frontal lobe}\el
            \gt{midline shifted to the left}\el \gt{glioblastoma}. \\
        \midrule
        CheXagent~\badmark
          & There is \hl{no evidence of intracranial hemorrhage or mass effect}.
            \emph{\textcolor{black!55}{(all of it)}} \\
        MAIRA-2~\badmark
          & \el the \hl{endotracheal tube terminating in the right mainstem
            bronchus}$^{\ddagger}$\el \\
        LLaVA-Rad~\badmark
          & \el \hl{opacities within the right middle and lower lobes, an infectious
            process}$^{\ddagger}$\el \\
        LLaVA-Med-1.5~\badmark
          & \el \hl{a large left frontal arachnoid cyst, a small left frontal
            subdural hematoma}\el \\
        \midrule
        Qwen2.5-VL (zero-shot)~\badmark$^{\dagger}$
          & \el \hl{a well-defined mass in the left parietal lobe}\el
            \hl{no mass effect or midline shift}\el \\
        $+$ Plain SFT~\badmark
          & \el the \hl{left frontal region}\el \hl{a slight rightward shift of the
            midline}\el \hl{meningioma}. \\
        $+$ \textbf{PerFact (ours)}~\okmark
          & \el the \gt{\textbf{right frontal lobe}}\el
            \gt{\textbf{leftward midline shift}}\el \gt{\textbf{Glioblastoma}}. \\
        \bottomrule
      \end{tabular}}%
  \sbox{\hcscan}{\includegraphics[trim=0.72bp 4.98bp 0.84bp 4.68bp, clip,
                                  width=\hcleftw]{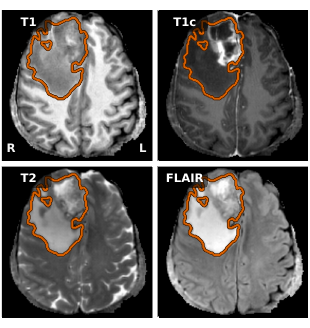}}%
  \sbox{\hcfact}{%
    \colorbox{factbg}{\parbox{\dimexpr\hcleftw-2\fboxsep\relax}{%
      \fontsize{6.2}{7.4}\selectfont\raggedright
      \textbf{Injected fact} (PerFact only):
      \textcolor{factblue}{\textbf{right frontal lobe}}; ring enhancement; edema;
      midline shift; ventricle compression;
      \textcolor{pathpurple}{glioblastoma}. \emph{No direction given.}}}}%
  \setlength{\hctabh}{\dimexpr\ht\hctab+\dp\hctab\relax}%
  \setlength{\hcslack}{0pt}%
  \sbox{\hcleft}{\vbox{\offinterlineskip\hsize\hcleftw
      \hbox{\usebox{\hcscan}}\vskip\hcslack\hbox{\usebox{\hcfact}}}}%
  \setlength{\hcslack}{\dimexpr\hctabh-\ht\hcleft-\dp\hcleft\relax}%
  \typeout{FIG2 ALIGN: table=\the\hctabh\space
           left=\the\dimexpr\ht\hcleft+\dp\hcleft\relax\space
           gap=\the\hcslack}%
  \ifdim\hcslack<0pt\relax\setlength{\hcslack}{0pt}\fi
  \sbox{\hcleft}{\vbox{\offinterlineskip\hsize\hcleftw
      \hbox{\usebox{\hcscan}}\vskip\hcslack\hbox{\usebox{\hcfact}}}}%
  \noindent\raisebox{-\height}{\usebox{\hcleft}}\hfill
  \raisebox{-\height}{\usebox{\hctab}}%
  \caption{\textbf{Injected facts correct a hallucinated location} (case
  BraTS-GLI-00599). Left: four sequences of one slice, radiological orientation,
  reference segmentation outlined, with the fact given to PerFact. Right: verbatim
  output, \dots\ marking an elision, green agreeing with the reference and orange
  contradicting it. The lower block varies only the injected text.
  $\dagger$\,A differential and a management plan rather than a report.
  $\ddagger$\,A thoracic structure.}
  \label{fig:hallucination}
\end{figure*}

Two routes dominate attempts to improve such systems, and we find that neither is where the
gain lies. The first is the model, either a larger backbone or pretraining on medical data:
six released medical and radiology vision-language models degrade sharply zero-shot on brain
MRI, the chest-specialized ones producing thoracic sentences for a brain study, and across
five backbones fine-tuned identically, from 3B to 32B parameters in three model families,
clinical entity F1 score varies by only 0.015 once the same facts are injected. The second is retrieval, conditioning on
similar prior reports or on curated knowledge: prior reports supply the form of a report
rather than its content, and retrieved knowledge is no better than plain fine-tuning. Neither
supplies what the generator is missing, a reliable statement of what is in the images.

What determines report quality is instead the information supplied to the model. We delegate perception to upstream 3D models, running segmentation and classification on the volumes, serialize their outputs into a compact \emph{fact sentence}, and leave the vision-language model the task it performs reliably: rendering structured facts as fluent clinical prose. We call this \textbf{PerFact}, for perception-derived facts. Holding backbone, split, target reports, adaptation recipe and decoding fixed and varying \emph{only} what is injected into the prompt, clinical entity F1 score increases from 0.644 with images alone to 0.775 with predicted facts, against an oracle ceiling of 0.879. That span of 0.235 is more than an order of magnitude beyond the 0.015 spanned by changing the model. Our contributions are summarized as follows:
\begin{itemize}
\item We present a controlled study of grounding on 3D brain MRI, over four diseases and three image contrasts, for both report generation and closed-ended Visual Question Answering (VQA), varying only the injected auxiliary information.
\item We show that model choice has minimal impact: report F1 score spans 0.015 across five backbones fine-tuned identically on the same facts, while six zero-shot medical and radiology vision-language models, most of all the chest radiology report generators, fail on brain MRI.
\item We show that the injected information has the greatest impact: F1 score spans 0.235 across the grounding conditions; predicted facts outperform retrieval, retrieval is redundant once facts are present, and predicted facts recover about half the oracle gain.
\item We identify where further effort is productive: no retrieval design survived reseeding, and the residual oracle gap is set by the granularity of the fact schema rather than by the generator.
\end{itemize}

\section{Related Work}

\paragraph{Radiology report generation:}
Radiology report generation grew out of image captioning \citep{showtell} and matured on chest radiographs, first with encoder-decoder and memory-driven architectures \citep{jing2018automatic,r2gen,r2gencmn} and more recently with report-specialized vision-language models such as MAIRA-2, LLaVA-Rad, and CheXagent \citep{maira2,llavarad,chexagent}. Benchmarks \citep{mimiccxr,iuxray}, entity extractors \citep{radgraph}, and clinical-efficacy metrics \citep{chexbert} were all designed around that modality. Volumetric work is thinner and differently shaped: the 3D radiology models that exist mostly target general-purpose captioning \citep{radfm}, while neuroimaging report generation has been approached either from a single disease family \citep{autorgbrain} or by pretraining on one health system's uncurated archive of millions of volumes \citep{neurovfm}. Neither settles which grounding signal drives report quality at ordinary corpus sizes, and we test the transfer of chest-trained systems rather than assuming it.

\paragraph{Grounding generation in facts and in retrieval:}
A recurring remedy for hallucination is to condition generation on structured intermediate representations: detected attributes \citep{promptmrg}, anatomical graphs \citep{kgrrg}, or knowledge injected from a curated ontology \citep{kiut,ppked}. Retrieval \citep{rag} is the other common source of context, and medical variants retrieve either similar prior cases or curated domain knowledge \citep{mmedrag,rule}. Both families usually compare one conditioned model against an unconditioned baseline, and their interaction is unexamined because the two are deployed in isolation. We contribute not a new conditioning mechanism but a controlled measurement of what each signal is worth against the alternatives.

\section{Method}

\begin{figure*}[t]
  \centering
  \includegraphics[width=0.98\textwidth]{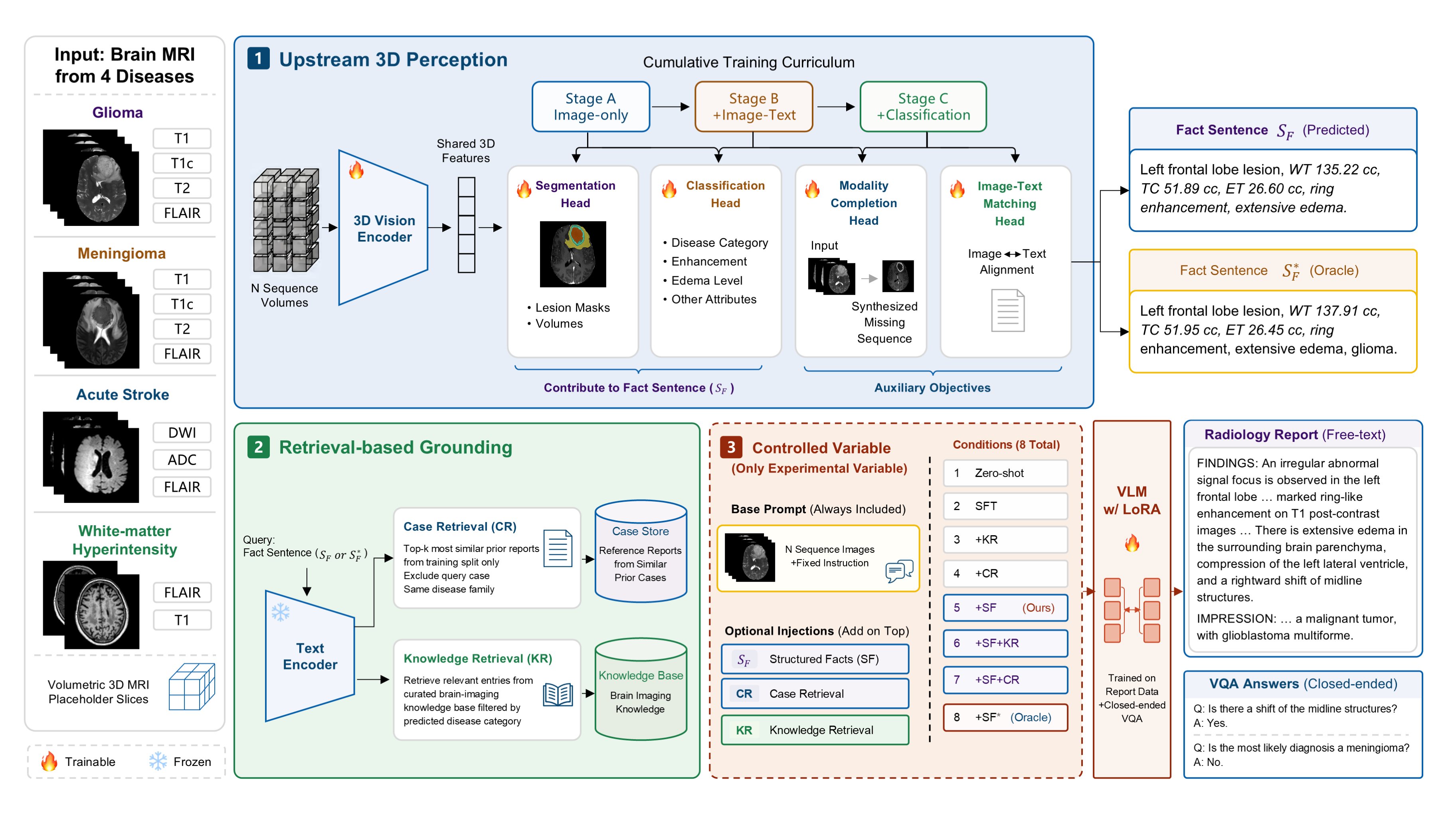}
  \caption{Overview of \textbf{PerFact}. Upstream 3D perception (top) turns a multi-contrast brain MRI study into a structured \emph{fact sentence} (SF); the same sentence is the query for retrieval over training-split reports (CR) or a knowledge base (KR) (bottom). The shared encoder drawn here is the tumor multi-task network; stroke and white-matter hyperintensity each use a standalone binary segmenter instead. The fact sentence, generated report, and VQA pairs are verbatim system output for one glioma case.}
  \label{fig:overview}
\end{figure*}

\subsection{Problem Formulation}
Let the dataset be $\mathcal{D} = \{(X_i, y_i, \mathcal{Q}_i)\}_{i=1}^{N}$, where $X_i = \{x_i^{(m)}\}_{m=1}^{M_i}$ is a brain MRI study with a variable number of co-registered 3D volumes with different image contrasts, $y_i$ is the target radiology report, and $\mathcal{Q}_i = \{(q_{i,j}, a_{i,j})\}_{j=1}^{J_i}$ is a set of closed-ended VQA pairs. Given $X_i$, the model must generate a report $\hat{y}_i$ and, when VQA supervision is included, predict the correct answer tokens for each question. We condition generation on an auxiliary grounding signal $\mathcal{G}_i$ and model the report distribution as $p_{\theta}(y_i \mid X_i, \mathcal{G}_i)$. The central question is which choice of $\mathcal{G}_i$ best improves $p_{\theta}$ under a fixed backbone, split, target report set, and decoding strategy. We write $d_i$ for the disease family of case $i$, $\mathcal{T}$ for the training split, and $\mathcal{B}$ for the curated neuroimaging knowledge base.

\subsection{PerFact Pipeline}
PerFact decomposes the problem into two stages (Figure~\ref{fig:overview}). First, upstream 3D perception maps the study to a structured latent summary. Second, that summary is serialized into a short fact sentence and used as prompt-level grounding for a vision-language generator. We write the upstream predictor as $z_i = h_{\phi}(X_i)$, where $z_i$ collects segmentation masks, lesion attributes, and disease labels predicted from the volumes. A deterministic serializer $\psi(\cdot)$ then converts these outputs into a fact sentence $s_i = \psi(z_i)$. This sentence is the factual interface between the perception model and the report generator. When ground truth is available for analysis, we also define an oracle sentence $s_i^* = \psi(z_i^*)$, which upper bounds the effect of conditioning without requiring inference-time annotation.

\subsection{Upstream 3D Perception}
Perception is trained on the full 3D study rather than on the 2D slices consumed by the generator. The upstream network is a multi-task model trained on four pre-tasks: lesion segmentation, missing-modality completion, image-text matching, and attribute classification. If we denote the corresponding losses by $\mathcal{L}_{\mathrm{seg}}$, $\mathcal{L}_{\mathrm{cmp}}$, $\mathcal{L}_{\mathrm{itm}}$, and $\mathcal{L}_{\mathrm{cls}}$, the upstream objective can be written in the generic weighted form
\begin{equation}
\mathcal{L}_{\mathrm{up}} = \lambda_{\mathrm{seg}}\mathcal{L}_{\mathrm{seg}} + \lambda_{\mathrm{cmp}}\mathcal{L}_{\mathrm{cmp}} + \lambda_{\mathrm{itm}}\mathcal{L}_{\mathrm{itm}} + \lambda_{\mathrm{cls}}\mathcal{L}_{\mathrm{cls}},
\end{equation}
where the $\lambda$ terms control the relative contribution of each pre-task. Training follows a three-stage curriculum $\mathcal{S}_1 \rightarrow \mathcal{S}_2 \rightarrow \mathcal{S}_3$, where $\mathcal{S}_1$ is image-only pretraining, $\mathcal{S}_2$ adds image-text alignment, and $\mathcal{S}_3$ adds classification supervision. We train each stage from scratch rather than warm-starting it from the previous stage, so that stage effects are not confounded with extra optimization. For non-tumor datasets, whose masks are binary rather than multi-region, we use a separate binary lesion segmenter per dataset, which avoids forcing the tumor label scheme onto stroke and white-matter disease \citep{segresnet}.

The serializer $\psi$ extracts disease-relevant fields from $z_i$, including laterality, lesion burden, lesion volume, lobe, enhancement pattern, and predicted disease category. These grouped attributes, $f_i = \bigl[f_i^{\mathrm{loc}}, f_i^{\mathrm{vol}}, f_i^{\mathrm{attr}}, f_i^{\mathrm{cat}}\bigr]$, are concatenated into a compact natural-language sentence $s_i = \psi(z_i)$. Coarse fields such as side, volume band, and burden are shared across diseases, while fine-grained fields such as multi-region volumes and enhancement pattern are only produced when the annotation supports them, which keeps the injected prompt human-readable.

\subsection{Retrieval and Prompt Construction}
The fact sentence is also the retrieval query. Let $e(\cdot)$ be the frozen text encoder learned by the image-text matching pre-task, and let $u_i = e(s_i)$ be the query embedding for case $i$. We use two retrieval sources.

For case retrieval, we define the candidate pool as the training split $\mathcal{T}$ and retrieve the top-$k$ reports within the same disease family. With cosine similarity $\mathrm{sim}(i,j) = \cos(u_i, u_j)$, the retrieval set is
\begin{equation}
\mathcal{R}_{\mathrm{CR}}(i) = \mathrm{TopK}_{j \in \mathcal{T},\, d_j = d_i} \mathrm{sim}(i,j),
\end{equation}
with the query case excluded from its own neighborhood. Retrieved reports are wrapped as reference text from similar prior cases rather than from the patient under study. For knowledge retrieval, we query a curated neuroimaging knowledge base $\mathcal{B}$ using the same sentence embedding:
\begin{equation}
\mathcal{R}_{\mathrm{KR}}(i) = \mathrm{TopK}_{b \in \mathcal{B}} \cos(u_i, e(b)).
\end{equation}
Retrieval is thus aligned with the structured fact representation rather than an image embedding or a sparse lexical match.

The final prompt concatenates the image tokens, a fixed instruction, and an optional grounding block, $P_i = [\mathrm{IMG}(X_i);\ \mathrm{INST};\ \mathcal{G}_i]$. The grounding block takes one of seven forms:
\begin{equation}
\mathcal{G}_i \in \left\{
\begin{array}{l}
\emptyset,\ \mathcal{R}_{\mathrm{KR}}(i),\ \mathcal{R}_{\mathrm{CR}}(i), \\
s_i,\ s_i^*,\ s_i \oplus \mathcal{R}_{\mathrm{KR}}(i),\ s_i \oplus \mathcal{R}_{\mathrm{CR}}(i)
\end{array}
\right\}.
\end{equation}
where $\oplus$ denotes concatenation. The \textbf{PerFact} condition is the simplest non-empty case, namely $\mathcal{G}_i = s_i$.

\subsection{Training Objective}
We adapt a general-purpose vision-language backbone with Low-Rank Adaptation (LoRA) \citep{lora} and keep the base model frozen. Let $\theta$ denote the backbone parameters and $\Delta\theta$ the LoRA updates. Report generation and closed-ended VQA are trained with a mixture objective $\mathcal{L} = (1-\rho)\,\mathcal{L}_{\mathrm{rep}} + \rho\,\mathcal{L}_{\mathrm{vqa}}$, where $\rho = 0.2$ in the main setting. The report loss is the standard autoregressive negative log-likelihood,
\begin{equation}
\mathcal{L}_{\mathrm{rep}} = - \sum_{i=1}^{N} \log p_{\theta+\Delta\theta}(y_i \mid X_i, P_i),
\end{equation}
and the VQA loss is the same token-level cross-entropy applied to the answer string
\begin{equation}
\mathcal{L}_{\mathrm{vqa}} = - \sum_{i=1}^{N}\sum_{j=1}^{J_i} \log p_{\theta+\Delta\theta}(a_{i,j} \mid X_i, q_{i,j}, P_i).
\end{equation}
Closed-ended VQA is scored without free-form generation: we read out the yes/no token mass at the answer position, normalise it, and optimize the model to place the higher mass on the correct answer.

\subsection{Evaluation}
For report generation, we report the clinical entity (CE) metrics, precision, recall and F1 score over the entities a report names, alongside the natural language generation (NLG) metrics BLEU \citep{papineni2002bleu}, METEOR \citep{banerjee2005meteor}, and ROUGE-L \citep{lin2004rouge}. Let $\mathrm{CE}(\cdot)$ denote the clinical entity extractor applied to generated and reference reports; then
\begin{equation}
\begin{array}{ll}
\mathrm{Prec} = \frac{|\mathrm{CE}(\hat{y}_i) \cap \mathrm{CE}(y_i)|}{|\mathrm{CE}(\hat{y}_i)|}, &
\mathrm{Rec} = \frac{|\mathrm{CE}(\hat{y}_i) \cap \mathrm{CE}(y_i)|}{|\mathrm{CE}(y_i)|},
\end{array}
\end{equation}
with F1 computed in the usual way from precision and recall. For VQA, we threshold the normalized yes-probability at 0.5 to compute accuracy and F1 score, and report the area under the receiver operating characteristic curve (AUROC) over the same score. Upstream perception is evaluated with Dice for segmentation and per-attribute AUC for classification.

\begin{table*}[!t]
  \centering
  %
  %
  \caption{\textbf{Grounding source is the lever} (report $n{=}130$, VQA $n{=}1549$).
  Expands the Qwen2.5-VL block of Table~\ref{tab:mainreport} across eight grounding
  conditions: $+$KR retrieved knowledge, $+$CR retrieved prior case reports, $+$SF
  predicted facts, $+$SF$^*$ ground-truth facts; only the injected information differs.
  Every cell is a single run, read against the seed band of the Experimental Setup.
  \textbf{Best} and \underline{second best} among deployable methods; $+$SF$^*$ is an
  upper bound, not a competitor.}
  \label{tab:main}
  \setlength{\tabcolsep}{3pt}
  \begin{tabular}{llccccccccc}
    \toprule
    \multirow{2}{*}{Method} & \multirow{2}{*}{Injected grounding}
      & \multicolumn{3}{c}{CE metrics} & \multicolumn{3}{c}{NLG metrics}
      & \multicolumn{3}{c}{Closed-ended VQA} \\
    \cmidrule(lr){3-5}\cmidrule(lr){6-8}\cmidrule(lr){9-11}
    & & PREC & REC & F-1 & BLEU-4 & METEOR & ROUGE-L & Acc & F-1 & AUROC \\
    \midrule
    Zero-shot & none, no adaptation & 0.541 & 0.480 & 0.509 & 2.68 & 20.32 & 15.52
      & 0.567 & 0.352 & 0.530 \\
    SFT       & none                & 0.671 & 0.619 & 0.644 & 25.83 & 43.71 & 49.85
      & 0.650 & \underline{0.567} & 0.697 \\
    \midrule
    $+$KR     & knowledge           & 0.682 & 0.601 & 0.639 & 25.98 & 43.96 & 51.31
      & 0.644 & \textbf{0.583} & 0.701 \\
    $+$CR     & prior case reports  & 0.701 & 0.736 & 0.718 & 28.15 & 46.70 & 51.68
      & \underline{0.683} & 0.488 & 0.701 \\
    \midrule
    \textbf{$+$SF (PerFact)} & \textbf{predicted facts}
      & \textbf{0.771} & \underline{0.779} & \underline{0.775}
      & \textbf{29.52} & \underline{47.30} & \textbf{53.55}
      & \textbf{0.693} & 0.535 & \textbf{0.720} \\
    $+$SF\,$+$KR & facts $+$ knowledge
      & \underline{0.763} & \textbf{0.791} & \textbf{0.777}
      & \underline{29.45} & \textbf{47.31} & \textbf{53.55}
      & 0.676 & 0.465 & \underline{0.711} \\
    $+$SF\,$+$CR & facts $+$ case reports
      & 0.766 & 0.754 & 0.760 & 29.20 & 46.85 & 52.63
      & 0.674 & 0.409 & 0.698 \\
    \midrule
    \emph{$+$SF$^*$} & \emph{oracle facts}
      & \emph{0.912} & \emph{0.849} & \emph{0.879}
      & \emph{34.28} & \emph{51.90} & \emph{56.81}
      & \emph{0.655} & \emph{0.546} & \emph{0.666} \\
    \bottomrule
  \end{tabular}
\end{table*}

\section{Experimental Setup}

\subsection{Data}
We build a unified multi-disease brain MRI corpus by reusing the radiology reports released with RadGenome-Brain MRI \citep{autorgbrain} across four public imaging sources: glioma (230 reported cases, T1-weighted/T1-weighted with contrast/T2-weighted/fluid attenuated inversion recovery [FLAIR]) \citep{brats2021}, meningioma (230, same sequences) \citep{bratsmen}, acute stroke (250, diffusion weighted imaging/apparent diffusion coefficient/FLAIR) \citep{isles22}, and white-matter hyperintensity (60 annotated training cases, FLAIR/T1-weighted) \citep{wmh}. The corpus spans four diseases and three image contrasts; the report test set contains 130 cases and the closed-ended VQA bank 1{,}549 questions. Because the sequence set differs by disease, the prompt carries a variable number of images and names the sequences present rather than padding to a fixed set.

We follow the official split where one exists and otherwise generate a 70/15/15 per-dataset split with seed 42, recorded in the released manifest. Cases are filtered for image-text quality per dataset rather than globally, so that a domain gap between datasets cannot silently remove one disease.

\paragraph{An external cohort without reference reports:}
We intended report-level external validation on a post-operative glioma cohort \citep{ucsf}, but its 298 studies provide segmentation labels and a clinical table with \textbf{no free-text reports}. We report this rather than substituting templated text, and confine external evidence to the upstream perception task.

\subsection{Implementation}
We use Qwen2.5-VL-7B-Instruct as the main backbone, and compare Qwen2.5-VL 3B/32B, LLaVA-1.5-7B, and InternVL3-8B under the same recipe. Fine-tuning uses LoRA \citep{lora} with $r{=}16$, $\alpha{=}32$, and all linear targets, trained with lr $1\mathrm{e}{-4}$, cosine decay, 0.05 warmup, and 3 epochs. Runs use bf16 precision, gradient checkpointing, batch size 1 with accumulation 8, a 4096--6144 token context, and \texttt{image\_max\_pixels} 262{,}144. We decode greedily throughout, and training runs on a single NVIDIA GH200 96GB GPU under aarch64 Linux, with LLaMA-Factory \citep{llamafactory}, CUDA 12.6, Python 3.11.7 and Transformers 4.49--4.53; upstream perception is implemented in PyTorch with MONAI \citep{monai}. Early runs used sampling, which altered F1 score by up to 0.03 in either direction; all reported results come from the greedy rerun.

Retrieval queries a frozen biomedical image-text encoder \citep{biomedclip} with the fact sentence. Closed-ended VQA is scored by the next-token yes/no probability mass rather than by generation, following the protocol of recent medical RAG work \citep{mmedrag}, so that accuracy, F1 score, and AUROC remain comparable to published radiology-VQA numbers.

\subsection{Measuring the Noise Band}
Retraining fixed configurations under training seeds 42, 1234 and 2024, report clinical entity F1 score has a standard deviation near 0.003 (0.001--0.005 across backbones) and VQA AUROC near 0.010, giving conservative bands of $\pm0.01$ and $\pm0.02$--0.03. Every comparison below is read against these bands, and differences inside them are reported as ties.

\subsection{Comparisons}
We vary three axes, each changing only one factor. \textbf{(i) Grounding source}: supervised fine-tuning on images alone (SFT) / $+$KR / $+$CR / $+$SF / $+$SF\,$+$CR / $+$SF$^*$; \textbf{(ii) backbone}: Qwen2.5-VL at 3B, 7B, and 32B \citep{qwen25vl}, LLaVA-1.5-7B \citep{llava15}, and InternVL3-8B \citep{internvl3}, all under fixed SF grounding; \textbf{(iii) domain pretraining}: six medical and radiology vision-language models evaluated zero-shot on the same test set and the same metric suite, namely MAIRA-2 \citep{maira2}, LLaVA-Rad \citep{llavarad}, CheXagent \citep{chexagent}, LLaVA-Med \citep{llavamed}, HuatuoGPT-Vision \citep{huatuogptvision}, and RadFM \citep{radfm}. Single-image models receive a $2\times2$ montage of the sequences; RadFM, the only natively volumetric model, receives the volumes.


\begin{table*}[!t]
  \centering
  \caption{\textbf{Fine-tuned backbones under identical PerFact grounding}, differing only
  in the backbone; every column is a mean over the same three seeds. \textbf{Best}
  and \underline{second best} in each column.}
  \label{tab:backbones}
  \setlength{\tabcolsep}{4pt}
  \begin{tabular}{llcccccccccc}
    \toprule
    \multirow{2}{*}{Method} & \multirow{2}{*}{LLM} & \multirow{2}{*}{Size}
      & \multicolumn{3}{c}{CE metrics} & \multicolumn{3}{c}{NLG metrics}
      & \multicolumn{3}{c}{Closed-ended VQA} \\
    \cmidrule(lr){4-6}\cmidrule(lr){7-9}\cmidrule(lr){10-12}
    & & & PREC & REC & F-1 & BLEU-4 & METEOR & ROUGE-L & Acc & F-1 & AUROC \\
    \midrule
    Qwen2.5-VL   & Qwen2.5  & 3B   & \underline{0.762} & 0.782 & 0.772 & 29.67 & 47.46 & \underline{53.86} & 0.618 & 0.570 & 0.683 \\
    Qwen2.5-VL   & Qwen2.5  & 7B   & 0.759 & \underline{0.793} & \underline{0.775} & \textbf{30.42} & \textbf{48.63} & \textbf{54.14} & 0.653 & 0.564 & 0.702 \\
    Qwen2.5-VL   & Qwen2.5  & 32B  & \textbf{0.766} & 0.782 & 0.774 & 29.90 & 47.75 & 53.60 & 0.640 & \underline{0.582} & \textbf{0.715} \\
    LLaVA-1.5    & Vicuna   & 7B   & 0.750 & 0.786 & 0.767 & \underline{29.99} & 47.56 & 53.64 & \underline{0.662} & 0.494 & 0.685 \\
    InternVL3    & Qwen2.5  & 8B   & 0.754 & \textbf{0.811} & \textbf{0.782} & 29.80 & \underline{47.90} & 53.44 & \textbf{0.675} & \textbf{0.603} & \underline{0.714} \\
    \bottomrule
  \end{tabular}
\end{table*}
\section{Results}


\subsection{Effect of the Grounding Source}

Holding everything else fixed and varying only what is injected, clinical entity F1 score varied
over a range of 0.235 (Table~\ref{tab:main}). Three observations follow. First, facts outperform retrieval: SF reaches 0.775 against 0.718
for case retrieval, while knowledge retrieval (0.639) is no better than
plain fine-tuning (0.644); retrieval helps only when it supplies the \emph{form} of a report,
not as a source of domain knowledge. Second, retrieval is redundant once facts are
present: adding knowledge to SF is flat ($+0.002$) and adding case reports slightly hurts
($-0.015$), because the two sources overlap in content rather than accumulate, which is why
PerFact is defined as SF alone. Third, imperfect upstream perception remains worthwhile: SF
recovers 0.131 of the 0.235 available above SFT, more than half the oracle gain, without
ground-truth annotation at inference; the residual oracle gap is 0.104.

Retrieval's contribution is stylistic, the 0.074 that CR alone adds over SFT. Sweeping its
design choices (query, retriever, $k$, reranking, filtering to content absent from the facts)
changed nothing: one variant appeared to gain $+0.021$ but did not survive reseeding.

The gain is not only aggregate. The injected fact corrects all three errors in
Figure~\ref{fig:hallucination}, and the direction of midline shift is recovered even though
the fact records only that a shift is present.
%
%

\subsection{Effect of Model Choice}

\paragraph{Backbone choice does not matter.}
Fixing grounding to SF and swapping only the backbone, clinical entity F1 score spans
0.767--0.782 across five models from three families and an order of magnitude in scale, over
three seeds each (Table~\ref{tab:backbones}; the best are the two PerFact rows of
Table~\ref{tab:mainreport}). That 0.015 spread is wider than reseeding alone predicts (one-way
ANOVA over the fifteen runs, $F(4,10){=}3.60$, $p{=}0.046$), so the backbones are not
interchangeable: InternVL3 separates, and the remaining four span 0.008. Read on a single
scale, the three axes we can vary are 0.006 wide for reseeding, 0.015 for the backbone and
0.235 for what is injected. Scale alone (3B, 7B, 32B) is flat at 0.772, 0.775, 0.774.

One asymmetry: report generation is flat across backbones, but VQA shows a mild, monotone
scale effect within the Qwen family . Rendering
structured facts as prose is within reach of every model tried; answering a closed question
about the image is not.

\begin{figure}[!t]
  \centering
  \includegraphics[width=\columnwidth]{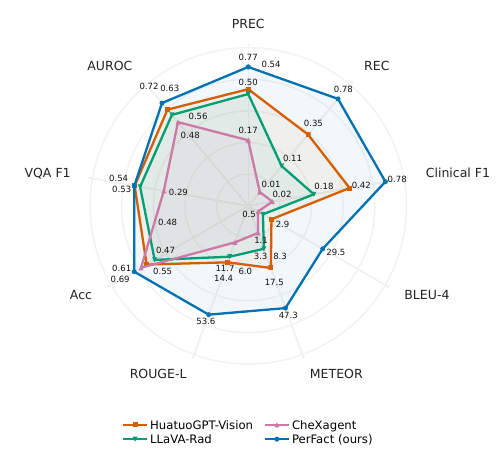}
  \caption{\textbf{PerFact against released models on all nine metrics}
  (zero-shot, brain MRI). The general medical model and the two chest radiology
  report generators all collapse, most sharply on the NLG axes, while PerFact
  traces a full profile. }
  \label{fig:radar_zeroshot}
\end{figure}

\begin{figure}[!tb]
  \centering
  \includegraphics{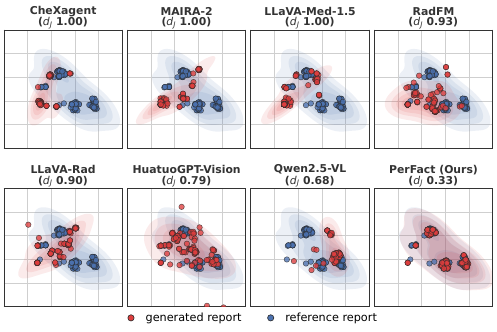}
  \caption{\textbf{t-SNE of clinical entity vectors against the reference reports}
  (brain MRI, $n{=}130$ each). All 1{,}170 vectors share one embedding, so the reference
  distribution is identical in every panel. Shading outlines each cloud and titles give the
  median Jaccard distance $d_J$. All panels but the last are zero-shot, and only PerFact's
  outline coincides with the reference's. The cluster left of center is the all-empty vector:
  86 of LLaVA-Med-1.5's reports name no recognized brain finding, against 3 of PerFact's.}
  \label{fig:tsne}
\end{figure}

\paragraph{Medical pretraining does not transfer.}
Table~\ref{tab:mainreport} evaluates six medical and radiology vision-language models
zero-shot. None approaches a fine-tuned system on any of the nine metrics
(Figure~\ref{fig:radar_zeroshot}), and performance orders cleanly by how far
the pretraining domain sits from brain MRI: general medical above biomedical and volumetric
radiology, with the chest radiology report generators at the bottom. The failure is
qualitative too, the chest-specialized models producing findings such as ``The heart size is
normal'' or references to PA/lateral views for a brain MRI study, and it takes more than one
form (Figure~\ref{fig:tsne}).
\begin{table*}[!t]
  \centering
  \caption{\textbf{Report generation and closed-ended VQA on brain MRI}
  ($n{=}130$ reports, $n{=}1549$ questions). Blocks: medical-pretrained VLMs, chest
  radiology report generators (both zero-shot), then ours with one un-adapted anchor per
  backbone; Table~\ref{tab:main} walks the intermediate conditions. LLaVA-Rad's report F-1
  is inflated by laterality words coinciding with brain entities; its recall of 0.108 is the
  honest figure. Answering no throughout scores 0.655 on the VQA bank, above every zero-shot
  model here, so AUROC is the column to read. Dashes: MAIRA-2 exposes no question-answering
  interface. \textbf{Best} and \underline{second best} among deployable rows; zero-shot and Qwen2.5-VL cells are single runs, the InternVL3 PerFact row a three-seed mean.}
  \label{tab:mainreport}
  \label{tab:mainvqa}
  \setlength{\tabcolsep}{2.4pt}
  \begin{tabular}{llcccccccccc}
    \toprule
    \multirow{2}{*}{Method} & \multirow{2}{*}{LLM} & \multirow{2}{*}{Size}
      & \multicolumn{3}{c}{CE metrics} & \multicolumn{3}{c}{NLG metrics}
      & \multicolumn{3}{c}{Closed-ended VQA} \\
    \cmidrule(lr){4-6}\cmidrule(lr){7-9}\cmidrule(lr){10-12}
    & & & PREC & REC & F-1 & BLEU-4 & METEOR & ROUGE-L & Acc & F-1 & AUROC \\
    \midrule
    HuatuoGPT-Vision                 & Qwen2    & 7B   & 0.544 & 0.346 & 0.423 & 2.87 & 17.46 & 14.43 & 0.550 & 0.532 & 0.631 \\
    RadFM                            & MedLLaMA & 14B  & 0.486 & 0.122 & 0.195 & 2.66 & 11.05 & 15.88 & 0.566 & 0.106 & 0.438 \\
    LLaVA-Med-1.5                    & Mistral  & 7B   & 0.503 & 0.090 & 0.153 & 0.70 & 8.13 & 11.84 & 0.394 & 0.525 & 0.547 \\
    \midrule
    LLaVA-Rad                        & Vicuna   & 7B   & 0.500 & 0.108 & 0.178 & 1.15 & 8.33 & 11.69 & 0.467 & 0.482 & 0.564 \\
    MAIRA-2                          & Vicuna   & 6.9B & 0.455 & 0.041 & 0.075 & 1.12 & 7.79 & 12.39 & -- & -- & -- \\
    CheXagent                        & Phi-2    & 8B   & 0.172 & 0.013 & 0.024 & 0.52 & 3.31 & 6.04 & 0.612 & 0.294 & 0.478 \\
    \midrule
    Qwen2.5-VL (zero-shot)           & Qwen2.5  & 7B   & 0.541 & 0.480 & 0.509 & 2.68 & 20.32 & 15.52 & 0.567 & 0.352 & 0.530 \\
    \textbf{PerFact (ours)}          &          &      & \textbf{0.771} & \underline{0.779} & \underline{0.775} & \underline{29.52} & \underline{47.30} & \textbf{53.55} & \textbf{0.693} & \underline{0.535} & \textbf{0.720} \\
    \cmidrule(l){1-12}
    InternVL3 (zero-shot)            & Qwen2.5  & 8B   & 0.499 & 0.455 & 0.476 & 2.05 & 20.57 & 17.06 & 0.522 & 0.467 & 0.576 \\
    \textbf{PerFact (ours)}          &          &      & \underline{0.754} & \textbf{0.811} & \textbf{0.782} & \textbf{29.80} & \textbf{47.90} & \underline{53.44} & \underline{0.675} & \textbf{0.603} & \underline{0.714} \\
    \bottomrule
  \end{tabular}
\end{table*}

%
%
%
%
%

The same ordering holds for closed-ended VQA (Table~\ref{tab:mainvqa}): the best zero-shot
model reaches AUROC 0.631, below PerFact's 0.720, and two fall \emph{below} chance. MAIRA-2
is not applicable rather than a number: it exposes only report generation and phrase
grounding, so a forced forward pass scores the first token of a report, not an answer.

\subsection{Upstream Perception}

The upstream models the facts depend on are competent in their own right. On the tumor
subset, adding image-text alignment to the multi-task network helps most: mean attribute AUC
over 17 attributes rises from 0.758 to 0.829 and tumor Dice from 0.755/0.746/0.731 to
0.764/0.767/0.767 (enhancing/core/whole), after which classification saturates and matching
recall at 10 rises from 0.07 to 0.48. Per-dataset segmenters reach Dice 0.697 on stroke and
0.756 on the official 110-case white-matter split \citep{wmh}, whose two unseen scanners
cost only 0.015 Dice. The white-matter segmenter was the one that most repaid attention:
retraining it at 500 epochs with best-validation checkpointing lifted its Dice from 0.592 to
0.750, a 27\% relative gain. See the Supplementary Material for details.

\subsection{Specificity to Report Generation}

Closed-ended VQA behaves differently. Across the six non-oracle conditions of Table~\ref{tab:main} the VQA AUROC range is
0.023, against 0.138 for clinical entity F1 score over those same conditions, so the
grounding lever is specific to the text the model writes. Answering ability instead requires having seen the answer format: without it
every condition sits near chance (AUROC 0.53--0.58), and the 20\% mixture we adopt lifts
AUROC to 0.696 at a cost to report F1 of $-0.002$, inside the noise band.

\subsection{Sources of the Residual Oracle Gap}

The residual $-0.104$ between SF and SF$^*$ in Table~\ref{tab:main} is not uniform; it tracks
how fine-grained the facts are (per-disease numbers in the supplement). Tumor cases carry the richest facts
(three region volumes, enhancement pattern, disease type) and lose the most when predicted,
F1 falling from 0.919 to 0.782. Stroke facts are coarser and lose almost nothing (0.782
against 0.762); white-matter facts are a single burden band and lose nothing at all. The
bottleneck is therefore the precision of what upstream can assert, not the generator: the
gap is widest exactly where the schema asks upstream for the most detail: three nested tumor
regions and an enhancement pattern against a single white-matter burden band. The
white-matter row rests on nine reported cases and is indicative rather than conclusive.
Whether a better
segmenter closes the gap is not something we measure here, since every end-to-end condition
reported above uses one fixed set of upstream checkpoints.

\section{Conclusion}

We asked what determines report quality on 3D multi-contrast brain MRI data, and answered it by
holding everything fixed except one variable at a time. Model choice is not the answer. The
injected information is: it varies clinical entity F1 score within a range of 0.235 against
the 0.015 spanned by the backbone, with
structured facts from upstream 3D perception outperforming retrieved prior reports, and
predicted facts recovering approximately half of the oracle headroom without inference-time
annotation. We instantiate this as
\textbf{PerFact}, the simplest configuration that carries the gain.

The remaining gap is upstream and structured: it scales with how much detail the fact schema
asks upstream for, widest for tumor cases and absent for white-matter cases. This directs
further work to the perception-language interface rather than to the generator. Two
constraints bound our claims: all reports come from a single annotation source, so the learned conventions are its own; and the generator consumes 2D slices, so a natively volumetric model might close part of the oracle gap. Both
are left to future work.

\clearpage
\label{sec:endcontent}  
\bibliography{references}

\end{document}


\maketitle

\section{Scope and Reading Guide}

This appendix introduces no new claims: every number is recomputed from the same stored
per-case predictions the paper reports, or specifies how those were produced. Nothing is
retuned, and no second evaluation protocol is introduced.

It runs specification first, at the detail needed to re-implement the method, then the
evidence behind claims the paper states in prose, then results resolved by disease and by
clinical entity. Three cautions apply throughout, and each is repeated where it bites.

\paragraph{Testbeds are not interchangeable.}
Every number in the main paper is measured on the unified four-disease corpus, $n{=}130$ for
report generation and $n{=}1{,}549$ for closed-ended visual question answering (VQA). Two
analyses here were measured on the earlier tumor-only corpus ($n{=}83$): the retrieval design
sweep of Section~H and the VQA mixture sweep. Both are labeled wherever they appear and are
not comparable to the four-disease numbers cell for cell.

\paragraph{Some rows rest on few cases.}
The white-matter hyperintensity split contributes nine reported cases, and the internal
segmentation test for that cohort is the same nine. Every white-matter number is indicative
rather than conclusive. The exception is the scanner-generalization result of Section~E,
measured on the official 110-case challenge test split.

\section{Data and Splits}

\subsection{Cohorts}

The corpus reuses the radiology reports released with RadGenome-Brain MRI
\citep{autorgbrain} over four public imaging sources (Table~\ref{tab:splits}). A fifth
RadGenome cohort, brain metastasis, was dropped before any experiment: its imaging is
archived off the distribution platform, and its 237 reports cannot be paired with volumes.

Sequences are not padded to a common set: glioma and meningioma carry T1-weighted,
T1-weighted with contrast, T2-weighted and fluid attenuated inversion recovery (FLAIR);
acute stroke, diffusion weighted imaging, the apparent diffusion coefficient map and FLAIR;
white-matter hyperintensity, FLAIR and T1-weighted.

\begin{table}[t]
\centering
\small
\setlength{\tabcolsep}{4pt}
\begin{tabular}{llrrrr}
\toprule
Cohort & Split source & Train & Val & Test & Total \\
\midrule
Glioma        & official  & 161 & 23 & 46 & 230 \\
Meningioma    & official  & 161 & 23 & 46 & 230 \\
Acute stroke  & seed 42   & 175 & 37 & 38 & 250 \\
White matter  & seed 42   &  42 &  9 &  9 &  60 \\
\midrule
Sourced       &           & 539 & 92 & 139 & 770 \\
Quality filter&           & $-14$ & $-1$ & $-9$ & $-24$ \\
\textbf{Used} &           & \textbf{525} & \textbf{91} & \textbf{130} & \textbf{746} \\
\bottomrule
\end{tabular}
\caption{\textbf{Every case, and where it went.} Cohorts without a released report-level
split are divided 70/15/15, seed 42. All reported numbers use these splits; the 130 test
cases are 83 tumor, 38 stroke and 9 white-matter.}
\label{tab:splits}
\end{table}

\subsection{Quality filter}

Cases are filtered per dataset, not globally, on image--report agreement scored by the
image--text matching pre-task. Twenty-four of the 770 fail, nine in the test split: 139 test
cases become the 130 every table reports. The filter runs once, before any experiment; every
condition uses the same 746 cases.

\subsection{Closed-ended VQA}

The VQA bank holds 1{,}549 yes/no questions over the 130 test cases. A deterministic script
builds them from the same case-level annotations as the fact schema of Section~D; the
archive ships it, not the questions. Questions take the form

\begin{quote}\small\ttfamily
Is the dominant lesion located in the left cerebral hemisphere? Answer with only yes or no.
\end{quote}

A question about a property the fact names is therefore answerable from the prompt alone.
The paper reads VQA as a control, not a second headline result: across the six non-oracle
conditions the range
is 0.023 in area under the receiver operating characteristic curve (AUROC), against 0.138 in
clinical entity F1 score.

\subsection{External cohort}

The post-operative glioma cohort \citep{ucsf} contributes 298 studies, one external test set
at the first of two timepoints. It has segmentation labels and a clinical table but no
free-text reports, supporting upstream evaluation only (Section~E).

\subsection{Availability}

All five sources are public and obtainable by any researcher; none is redistributed here.
The reports are MIT-licensed. The imaging is not: the stroke release requires the organizers'
written agreement, the white-matter release forbids redistributing the data or anything
derived from it, and the tumor and external cohorts come under a data-use agreement that
does not grant redistribution.

\section{Configuration}

Table~\ref{tab:config} lists every setting that affects a reported number. The generator's
optimizer, weight decay, gradient clipping and LoRA dropout are not set in any configuration
file and take the framework defaults. Context length is the only per-condition deviation:
6144 tokens for the two retrieval-augmented conditions, which carry three truncated reports,
and 4096 for the rest. The vision tower is not separately frozen; LoRA covers all linear
layers. The upstream network trains in fp32 with no mixed precision, unlike the generator;
its 3D convolutions are memory-bound at this patch size.

\begin{table*}[!t]
\centering
\small
\setlength{\tabcolsep}{3pt}
\begin{minipage}[t]{0.48\textwidth}
\centering
\begin{tabular}{ll}
\toprule
Setting & Value \\
\midrule
\multicolumn{2}{l}{\emph{Data}}\\
Corpus                     & 746 cases, four diseases \\
Report test set            & 130 cases \\
Question bank              & 1{,}549 yes/no questions \\
Split seed                 & 42 \\
Images per case            & one axial slice per sequence \\
\midrule
\multicolumn{2}{l}{\emph{Upstream multi-task network}}\\
Encoder widths             & $(32, 64, 128, 256)$, 3D \\
Embedding / classes        & 512-d / 24 attributes \\
Patch size                 & $96^3$ random crop \\
Objective                  & Dice, $L_1$, 0.2 InfoNCE, 0.3 BCE \\
Matching temperature       & 0.07 \\
Optimizer                  & AdamW, $2\times10^{-4}$, wd $10^{-5}$ \\
Schedule                   & none \\
Batch / epochs             & 4 / 30 per curriculum stage \\
Stage initialization       & from scratch, not warm-started \\
Precision                  & fp32, no mixed precision \\
\midrule
\multicolumn{2}{l}{\emph{Binary lesion segmenters}}\\
Architecture               & 3D residual, 16 filters $(1,2,2,4)$ \\
Loss / optimizer           & Dice / AdamW $10^{-4}$, wd $10^{-5}$ \\
Patch / batch              & $96^3$ / 2 \\
Epochs                     & 200 stroke, 500 white matter \\
Model selection            & best validation Dice \\
Augmentation               & random crop, random flip \\
Inference                  & sliding window \\
\bottomrule
\end{tabular}
\end{minipage}\hfill
\begin{minipage}[t]{0.48\textwidth}
\centering
\begin{tabular}{ll}
\toprule
Setting & Value \\
\midrule
\multicolumn{2}{l}{\emph{Retrieval}}\\
Encoder                    & frozen biomedical image--text \\
Query                      & the fact sentence \\
Retriever / $k$            & BM25 / 3 \\
Candidate pool             & training split, same disease \\
Exclusions                 & self, identical fact sentence \\
Truncation                 & 400 chars (cases), 420 (knowledge) \\
\midrule
\multicolumn{2}{l}{\emph{Report generator}}\\
Backbone                   & Qwen2.5-VL-7B-Instruct \\
Adaptation                 & LoRA, rank 16, $\alpha$ 32, all linear \\
LoRA dropout               & not set (framework default) \\
Learning rate              & $1\times10^{-4}$, cosine, warmup 0.05 \\
Optimizer / weight decay   & not set (framework default) \\
Epochs                     & 3 \\
Batch                      & 1 $\times$ 8 accumulation $=$ 8 \\
Precision                  & bf16, gradient checkpointing \\
Context length             & 4096; 6144 with retrieval \\
Image budget               & 262{,}144 pixels per image \\
Model selection            & final epoch \\
\midrule
\multicolumn{2}{l}{\emph{Decoding and scoring}}\\
Report decoding            & greedy, 512 new tokens \\
Question scoring           & yes/no token mass, no generation \\
Threshold                  & 0.5 for accuracy and F1 \\
Seeds                      & 42; 42/1234/2024 for backbones \\
\bottomrule
\end{tabular}
\end{minipage}
\caption{\textbf{Every setting that affects a reported number.} Read from the run
configurations. Entries marked ``not set'' are absent from those files and take the framework
default; they are unspecified, not guessed.}
\label{tab:config}
\end{table*}

\section{The Fact Schema and the Prompts}

\subsection{The instruction}

All conditions share one instruction: for a case with $M$ sequences, $M$ image tokens
followed by

\begin{quote}\small\ttfamily
These are the T1, T1c, T2 and FLAIR MRI sequences of a brain tumour patient. Write the
radiology Findings and Impression.
\end{quote}

Sequence names are those actually present, comma-joined with a final ``and''; the patient
descriptor is ``brain tumour patient'' for the two tumor cohorts and ``brain MRI patient''
otherwise. There is no system prompt: records carry a user and an assistant message, and any
system text comes from the backbone's stock template. There are no few-shot exemplars, no
length or style instruction, no output schema, and no statement that the images are single
axial slices.

The assistant target is the reference report, beginning \texttt{Findings:} with an
\texttt{Impression:} section.

\paragraph{Which slice reaches the model.}
One image per sequence, every sequence of a case cut at the \emph{same} axial index: where
the reference lesion mask is widest, after reorienting to a canonical axis order. Sequences
on another grid are resampled to the mask's first; intensities are windowed at the 0.5 and
99.5 percentiles and written as 8-bit images. A tumor case contributes four images, a stroke
case three, a white-matter case two. This is a
strong simplification -- one plane, one level, chosen by an annotation the model does not see
at inference -- and why the paper frames the generator as reading three of hundreds of
slices.

\subsection{The fact sentence}

The serializer emits one line, opened by a fixed prefix, fields joined by semicolons:

\begin{quote}\small\ttfamily
Auxiliary findings from upstream segmentation/classification models -> location: left
frontal lobe; whole-tumor volume 137.91 cm3 (enhancing 26.45 cm3, core 51.95 cm3); features:
ring-like enhancement, extensive peritumoral edema, midline shift; predicted category:
glioma, glioblastoma.
\end{quote}

Four field groups can appear. \emph{Location} is the predicted side, dropped when
\texttt{unknown} or \texttt{midline}, then the predicted lobes from frontal, temporal,
parietal, occipital, cerebellum and brainstem, joined with a slash; with neither the slot
reads \texttt{intra-axial}. \emph{Volume} appears only where three-region segmentation
exists, as whole-tumor volume in cm$^3$ with enhancing and core volumes in parentheses,
computed from the mask at 1\,mm isotropic spacing. \emph{Features} draws on a closed
vocabulary of eight strings: ring-like enhancement, enhancement present, necrosis, extensive
peritumoral edema, peritumoral edema, midline shift, lateral-ventricle compression, and
crosses midline / bilateral; enhancement and edema each have a graded and an ungraded form
and only one of each pair is emitted. \emph{Predicted category} is at most two disease
labels.

The volume triple and the enhancement pattern exist only for the tumor cohorts; stroke
contributes a lesion and its location, white-matter hyperintensity a burden level.

The oracle runs the same serializer over ground-truth masks and labels, so the two differ
only in their values.

\subsection{The retrieval blocks}

Case retrieval prepends a header and up to three numbered reports, each truncated to 400
characters at a word boundary:

\begin{quote}\small\ttfamily
Reference reports from similar prior cases (different patients, for grounding only; do not
copy verbatim):\\{}[1] Findings: \ldots{} Impression: \ldots
\end{quote}

Knowledge retrieval uses the same shape with a 420-character limit:

\begin{quote}\small\ttfamily
Reference medical knowledge (general radiology knowledge from medical textbooks, NOT
specific to this patient; use it to interpret the images, do not copy verbatim):
\end{quote}

Both are top-3.

\subsection{How the blocks compose}

Blocks are appended to the instruction, separated by blank lines:

\begin{center}\small\ttfamily
$\mathcal{G}=\emptyset$:\quad INST\\
$+$SF:\quad INST $\backslash$n$\backslash$n FACTS\\
$+$CR:\quad INST $\backslash$n$\backslash$n CASES\\
$+$SF$+$CR:\quad INST $\backslash$n$\backslash$n FACTS $\backslash$n$\backslash$n CASES
\end{center}

Ordering is fixed across conditions and splits; records are emitted in manifest order with
no shuffling.

\subsection{What cannot be shown here}

The retrieved-knowledge condition reads the MedRAG textbook corpus \citep{medrag}, a public
release of 125,847 pre-chunked passages from 18 English medical textbooks. We keep the
brain-imaging-relevant chunks with a two-tier keyword filter, roughly 2,100 of the 125,847,
and leave per-case precision to the dense and BM25 retrieval downstream. The corpus is not
copied into the code archive, because it is not ours to redistribute, but it is downloadable
by anyone and the filter that selects from it ships, so the condition is reproducible end to
end. What we do not reproduce here is the passage text itself, which is under copyright: the
wrapper template above is exact, but the three slots it fills cannot be quoted. That
condition is in any case one of the paper's negative results, no better than fine-tuning
alone.

The released models get their own instructions, each a single sentence of the same shape,
for example ``This is a multi-sequence brain MRI. Write the radiology report: Findings and
Impression.'' 

\section{Upstream 3D Perception}

\subsection{The tumor multi-task network}

A single 3D convolutional encoder, feature widths $(32, 64, 128, 256)$, feeds four heads
trained together: lesion segmentation, missing-modality completion, image--text matching over
a 512-d embedding, and classification over 24 candidate attributes. Inputs are four
co-registered sequences at $96^3$ random crops.

Eq.~(1) of the main paper instantiates as
$\mathcal{L}_{\mathrm{up}} = \mathcal{L}_{\mathrm{seg}} + \mathcal{L}_{\mathrm{cmp}}
+ 0.2\,\mathcal{L}_{\mathrm{itm}} + 0.3\,\mathcal{L}_{\mathrm{cls}}$, with Dice on the
segmentation head, $L_1$ for completion, InfoNCE at temperature $0.07$ for matching, and
binary cross-entropy (BCE) per attribute; optimization is as in Table~\ref{tab:config}.

The three curriculum stages are $\mathcal{S}_1$, image-only pretraining;
$\mathcal{S}_2$, which adds image--text alignment; and $\mathcal{S}_3$, which adds
classification supervision. Each runs 30 epochs and trains from scratch, not from the
preceding stage, so a stage effect cannot be longer optimization in disguise.

\begin{figure*}[!t]
  \centering
  \includegraphics[width=\textwidth]{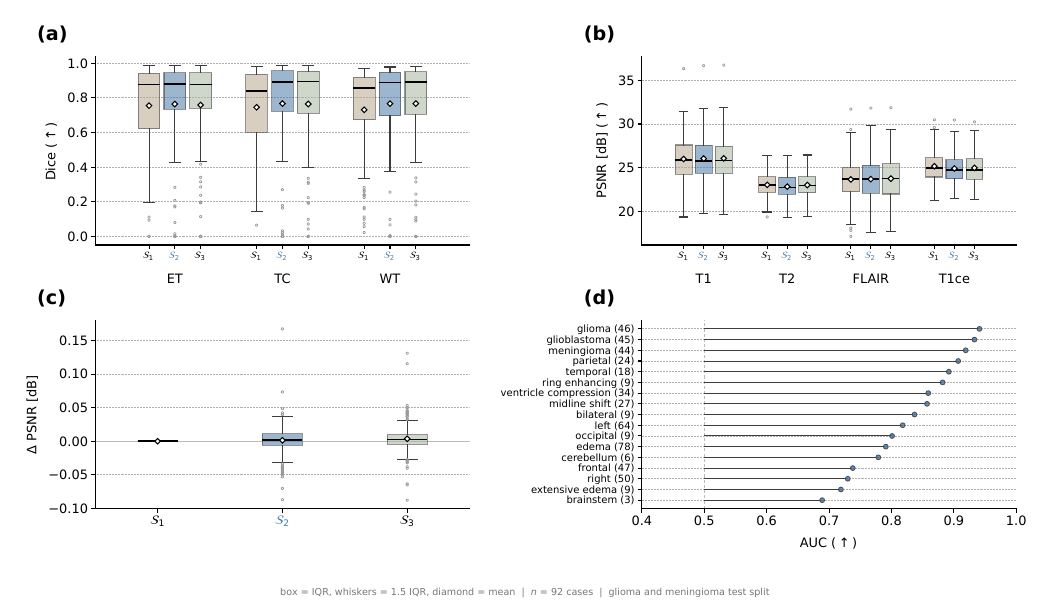}
  \caption{All four pre-tasks across the three curriculum stages, labeled by stage on the
  axis. $\mathcal{S}_2$ lifts segmentation (a) by shortening the lower tail, not by moving
  the typical case, so the medians barely differ. Completion (b) does not move, and neither
  does the image-text branch's effect on it (c), zero at $\mathcal{S}_1$ for want of a text
  branch. Per-attribute AUC (d) is at the deployed stage; six of the 17 rest on nine cases
  or fewer.}
  \label{fig:pretasks}
\end{figure*}

To three decimals: mean attribute AUC over the 17 scorable attributes runs 0.758, 0.829, 0.823 across the stages; Dice on the enhancing, core
and whole regions runs 0.755/0.746/0.731, 0.764/0.767/0.767, 0.759/0.765/0.767; completion is
flat at 24.48, 24.38 and 24.46\,dB, and removing the text branch changes it by under
0.01\,dB at every stage. The deployed checkpoint is $\mathcal{S}_2$.

\subsection{Image--text matching recall}

Contrastively fine-tuning a 512-d projection on the frozen biomedical encoder
\citep{biomedclip} lifts recall at 10 from 0.07 to 0.478 and at 1 from 0.009 to 0.130, and
drops the median rank from 89 to 12. The two are not a controlled pair: the
zero-shot row ranks each case against the whole manifest, the fine-tuned row within the
46-case test split, so the gain is an upper bound. Retrieval at inference uses the frozen
encoder rather than the fine-tuned projection.

\subsection{The two binary lesion segmenters}

Stroke and white-matter hyperintensity use a standalone binary segmenter each
\citep{segresnet} rather than the tumor label scheme, configured as in
Table~\ref{tab:config} and keeping the best-validation checkpoint.

On the internal splits, seeded at 42, stroke reaches Dice 0.697 at 200 epochs over 37 of 38
test cases; the 38th has an empty reference. The white-matter cohort is a
quarter the size, so 120 epochs left it undertrained at 0.592; 300 epochs reach 0.729 and 500
reach 0.750, with the best validation Dice of 0.790 at epoch 350. Those three white-matter
figures rest on nine cases and are indicative only. The same checkpoint scores 0.756 on the
official challenge split, which is the figure the main paper reports.

\subsection{Held-out evaluation on the official challenge test set}

Only the white-matter segmenter has a public held-out test set. Training and model selection
used 51 and 9 of the 60 official training cases; the 110 test cases were untouched until
final evaluation.

The benchmark's best public entry \citep{wmh} reaches 0.81; this is a generic architecture
trained on 51 cases inside a report-generation system, not a method tuned for the challenge.

Twenty of those 110 come from two scanners absent from training
(Table~\ref{tab:scanners}).

\begin{table}[t]
\centering
\small
\setlength{\tabcolsep}{4pt}
\begin{tabular}{llrr}
\toprule
Site and scanner & Seen in training & $n$ & Dice \\
\midrule
Utrecht, 3\,T Philips        & yes & 30 & 0.738 \\
Singapore, 3\,T Siemens      & yes & 30 & 0.775 \\
Amsterdam, 3\,T GE           & yes & 30 & 0.764 \\
Amsterdam, 1.5\,T GE         & \textbf{no} & 10 & 0.758 \\
Amsterdam, 3\,T Philips      & \textbf{no} & 10 & 0.730 \\
\midrule
Seen                         &     & 90 & 0.759 \\
Unseen                       &     & 20 & 0.744 \\
\bottomrule
\end{tabular}
\caption{\textbf{Two unseen scanners cost 0.015 Dice.} Official challenge test split, 110
held-out cases. The seen--unseen gap is smaller than the spread among the seen scanners.}
\label{tab:scanners}
\end{table}

\subsection{A second cohort, from another institution}

The tumor segmenter was applied unchanged to a post-operative glioma cohort from another
institution \citep{ucsf}: 298 studies, a scanner and protocol absent from training. Whole
tumor, the one region non-empty in every reference, scores 0.577.

Dice against an empty reference is zero by convention, not a measurement, and 111 of the 298
studies contain no enhancing component, so a cohort-wide mean measures composition as much as
quality. Table~\ref{tab:external} therefore groups by content; per-study Dice for all 298 is
in the archive.

\begin{table}[t]
\centering
\small
\setlength{\tabcolsep}{4pt}
\begin{tabular}{lrrrr}
\toprule
Studies & $n$ & Enh. & Whole & Core \\
\midrule
With an enhancing tumor              & 187 & 0.576 & 0.659 & 0.463 \\
\quad and no resection cavity        &  20 & \textbf{0.753} & \textbf{0.763} & 0.640 \\
\midrule
Internal test, for reference         &  92 & 0.759 & 0.767 & 0.765 \\
\bottomrule
\end{tabular}
\caption{\textbf{The segmenter transfers where the anatomy transfers.} Mean Dice on the
second cohort by what the reference contains. Where the anatomy matches training -- an
enhancing tumor, no resection cavity -- enhancing and whole-tumor Dice fall within 0.006 and
0.004 of the internal test set, untrained.}
\label{tab:external}
\end{table}

Upstream perception is thus evaluated twice outside its training data: a change of scanner
costs 0.015 Dice, a change of institution with the anatomy held comparable 0.006, and a
change in the anatomy itself more than either, tumor core falling furthest where a resection
cavity most disturbs it. Post-operative anatomy and studies with no enhancing component lie
outside what it covers here.

\section{Evaluation Metrics}

\subsection{The clinical entity (CE) extractor}

$\mathrm{CE}(\cdot)$ is rule-based, not learned: substring and regular-expression tests over
a fixed vocabulary of 24 labels, on lowercased text. It is not CheXbert, RadGraph, scispaCy
or any UMLS linker.

The 24 labels are \texttt{infarct}, \texttt{meningioma}, \texttt{metastasis},
\texttt{sulcal\_effacement}, \texttt{edema}, \texttt{necrotic}, \texttt{non\_enhancing},
\texttt{enhancing}, \texttt{brainstem}, \texttt{cerebellum}, \texttt{midline},
\texttt{ventricle\_compression}, \texttt{bilateral}, \texttt{right},
\texttt{midline\_shift}, \texttt{left}, \texttt{parietal}, \texttt{frontal},
\texttt{occipital}, \texttt{extensive\_edema}, \texttt{ring\_enhancing},
\texttt{temporal}, \texttt{glioma} and \texttt{glioblastoma}. Twenty-one occur in the
reference reports of the test split; metastasis, sulcal effacement and necrosis do not.

Five properties matter.

\paragraph{Presence, not count.} A label is set if any pattern fires anywhere in the report,
and contributes at most one true positive per case.

\paragraph{Laterality is an entity.} \texttt{left}, \texttt{right} and \texttt{bilateral} are
three of the 24, matched on word boundaries; Section~K treats laterality as content, not
style.

\paragraph{Substring matching, with three exceptions.} Most labels test a stem:
\texttt{necro} for necrosis and necrotic, \texttt{cerebell} for cerebellum and cerebellar,
\texttt{enhanc} for enhancing, enhancement and enhanced. Three rules are stricter:
\texttt{glioblastoma} implies \texttt{glioma}; \texttt{enhancing} is suppressed when
\texttt{non\_enhancing} fires; \texttt{extensive\_edema} and
\texttt{ventricle\_compression} require their two words within a bounded window in one
sentence.

\paragraph{Negation is handled for one label.} \texttt{midline\_shift} is set by a
positive pattern and vetoed by a negative one, so ``no midline shift'' does not set it. No
other label is negation-aware: ``no edema'' sets \texttt{edema}. Applied identically to
reference and generated report, this inflates agreement symmetrically; the metric compares
two rule-parsed vectors, not a report against a clinician.

\paragraph{The same rules define the attribute labels.} The upstream network's
classification targets come from this same function: the fact sentence and the metric share
one piece of code. 

Precision, recall and F1 score are \emph{micro}-averaged: true positives, false positives
and false negatives are pooled over all cases and all labels before the ratio, so frequent
entities outweigh rare ones. No macro average is computed. True negatives are never counted,
so correctly omitting an absent finding earns nothing.

\subsection{Natural language generation (NLG) metrics}

BLEU is corpus-level, from \texttt{sacrebleu} at cumulative orders one to four with
effective-order smoothing; the tables print BLEU-4. ROUGE-L and METEOR are sentence-level,
macro-averaged over case pairs. All three are multiplied by 100, so the tables print them on
0--100 while the clinical entity metrics stay on 0--1; the radar places the two groups on
one ring.

\subsection{Closed-ended visual question answering}

Nothing is generated: the model runs once over the prompt, reading the full-vocabulary
softmax at the final prompt position. The first-token probability mass of six affirmative
surface forms is summed into $y$, the six negative forms into $n$, and
$p_{\mathrm{yes}} = y/(y+n+\varepsilon)$. Accuracy and F1 score threshold
$p_{\mathrm{yes}}$ at 0.5 with ``yes'' as the positive class; AUROC uses the unthresholded
score. The read-out follows \citet{mmedrag} and lets a model score below chance rather than
at it.

MAIRA-2 exposes report generation and phrase
grounding but no interface for this read-out. The dash in the cross-model table is a missing
measurement, not a failure.

\subsection{Upstream metrics}

Dice is computed per case and averaged over cases, separately for the enhancing, core and
whole tumor regions of the challenge label scheme. Cases whose reference is empty for a
region are skipped where noted, counted as zero where the region is genuinely absent;
Section~E gives the convention per row, a distinction that dominates the external-cohort
result. Attribute classification is scored with AUROC per attribute, averaged over
attributes for which both classes are present in the test split.

\section{Seed Variation and Statistics}

\subsection{Two numbers for the same noise}

The main paper quotes reseeding noise twice, as a standard deviation near 0.003 and as
``0.006 wide''; these are one measurement read two ways. Retraining one fixed configuration
under seeds 42, 1234 and 2024 gives clinical entity F1 scores of 0.7766, 0.7818 and 0.7824:
standard deviation 0.0032, range 0.0058. The $\pm0.01$ band used throughout is roughly three
standard deviations.

\subsection{The fifteen runs behind the backbone comparison}

Table~\ref{tab:seeds} lists every run behind the five column means the paper reports.

\begin{table}[t]
\centering
\small
\setlength{\tabcolsep}{4pt}
\begin{tabular}{lccccc}
\toprule
Backbone & s42 & s1234 & s2024 & Mean & SD \\
\midrule
Qwen2.5-VL-3B  & 0.7736 & 0.7670 & 0.7746 & 0.7717 & 0.0041 \\
Qwen2.5-VL-7B  & 0.7751 & 0.7742 & 0.7768 & 0.7754 & 0.0013 \\
Qwen2.5-VL-32B & 0.7677 & 0.7802 & 0.7736 & 0.7738 & 0.0062 \\
LLaVA-1.5-7B   & 0.7628 & 0.7723 & 0.7667 & 0.7673 & 0.0048 \\
InternVL3-8B   & 0.7796 & 0.7770 & 0.7883 & 0.7817 & 0.0059 \\
\bottomrule
\end{tabular}
\caption{\textbf{Clinical entity F1 score for every backbone run.} Five backbones, three
seeds each, identical grounding, data and recipe. Spread of the five means 0.0144; pooled
within-backbone standard deviation 0.0048.}
\label{tab:seeds}
\end{table}

A one-way analysis of variance (ANOVA) over these fifteen runs gives $F(4,10) = 3.60$, $p = 0.046$: the five
backbones are not exchangeable. InternVL3-8B has the highest mean, 0.7817, and removing it
leaves the other four spanning 0.0081, inside the noise band.

Read on one scale, the three axes are 0.006 wide for reseeding, 0.0144 for the backbone and
0.235 for the injected grounding.

\subsection{How many runs stand behind each reported cell}

The grounding ladder is single runs at seed 42; its eight conditions differ by far more than
the noise band. The backbone comparison is three-seed means. The cross-model comparison mixes
the two: zero-shot and fine-tuned Qwen2.5-VL rows are single runs, the InternVL3 PerFact row
a three-seed mean, since zero-shot models are deterministic under greedy decoding. Upstream
perception is a single run per configuration; the segmentation and classification results in
Section~E carry no variance estimate.

\subsection{What is not tested}

Only the backbone comparison is tested statistically, and no correction for multiple
comparisons is applied: the
effects the paper argues from are between five and forty times the band. Differences that are
not are reported as ties.

\section{Ablations the Paper States in Prose}

\subsection{The retrieval design sweep}

The sweep was run on the \emph{tumor-only} corpus, $n{=}83$, so its values are not
comparable cell for cell with any other table
here. On that testbed the grounding ladder holds as it does on the four-disease one: 0.603
without grounding, 0.735 with retrieved cases, 0.759 with facts, 0.765 with both, and 0.911
with the oracle.

Four variants were run against the fact-plus-cases base at 0.765. Complement-only, which keeps
only retrieved text the fact sentence does not already carry, reached 0.785, $+0.020$;
knowledge-filtered, which selects by match against the knowledge base, 0.767, $+0.002$;
reranking, 0.755, $-0.009$; and replacing the fact sentence with an image embedding as the
retrieval key, 0.754, $-0.011$. Only the first is outside the noise band, and it is the
$+0.021$ the main paper reports as not surviving reseeding.

Complement-only is what the
redundancy finding predicts should help: if the two sources overlap rather than accumulate,
removing the overlap should recover the difference. It appeared to, by $+0.020$, then did
not reproduce.

How many reports to retrieve was calibrated the same way, varying $k$ at inference on one
fixed adapter so the arms differ only in the prompt: on the glioma and meningioma validation
split, $n{=}45$, clinical entity F1 runs 0.738 at $k{=}0$, then 0.815, 0.798 and 0.818 at
$k{=}1$, 2 and 3. The first retrieved report carries the gain and the next two do not.
$k{=}5$ is not a separate condition, because the same-disease pool never yields a fifth
report and its prompts are the $k{=}3$ prompts on all 45 cases.

Four image-text encoders over one corpus, scored as in \citet{mmedrag}, say why the retrieval
result reads the way it does. Exact-report recall at 10 runs 2.4, 2.4, 5.4 and 7.2 for CLIP,
PubMedCLIP, PMC-CLIP and BiomedCLIP; same-disease precision at 3 runs 65.4, 64.5, 77.0 and
67.6. Retrieval is therefore not supplying the right report often enough for that to be where
its value lies; what it reliably supplies is a report of the right disease, in the right form.
That is the sense in which the main paper calls retrieval's contribution stylistic, and it is
consistent with the first report carrying all of the gain and the second and third carrying
none. BiomedCLIP is the encoder deployed, though PMC-CLIP retrieves the same disease more
often.

\subsection{The curriculum}

Section~E gives the three-stage ablation (Figure~\ref{fig:pretasks}): image-text alignment
moves both heads, the classification objective neither.

\subsection{Mixing question answering into training}

Report-only training leaves closed-ended question answering near chance for every grounding
condition, at 0.53--0.58 area under the curve. Mixing questions into training at 20\% lifts it
to 0.696 at a cost of $-0.002$ in report F1 score, inside the noise band, so we adopt it for
every reported condition. A 40\% mixture lifts VQA further at a larger cost to report F1
score; 20\% is the smallest that clears the chance band.

The seven-condition version exists only on the tumor-only corpus and ships in the code archive,
not beside the four-disease numbers (Section~A).

\subsection{Decoding}

Early runs used the training framework's sampling default rather than greedy decoding. The
difference moved clinical entity F1 score by up to 0.03 in either direction, three times the
reseeding band and enough to reorder adjacent conditions. No reported comparison mixes the two. The sampling runs were discarded
rather than analyzed.

\section{Backbones and Released Models}

\subsection{The grounding ladder on all three backbones}

Table~\ref{tab:ladder_backbones} repeats \mainTabLadder{} unchanged on the other two
fine-tuned backbones, so the central comparison rests on three models rather than one.

\begin{table*}[!t]
  \centering
  \caption{\textbf{The grounding ladder on all three fine-tuned backbones.} The eight conditions and nine metrics of \mainTabLadder{}, re-run unchanged; report $n{=}130$, questions $n{=}1{,}549$. Qwen2.5-VL-7B is \mainTabLadder{} itself and is not repeated here; the generator checks the two against each other on all 72 cells before writing. The two LLaVA-1.5 question-answering F1 cells at 0.000 and 0.040 are Section~F's calibration effect at its extreme: at the fixed 0.5 threshold both answer \emph{no} to nearly every question, emptying the positive class while the ranking holds at 0.637 and 0.667 area under the curve; at their own best threshold, 0.532 and 0.554. \textbf{Best} and \underline{second best} within each block among deployable methods; the oracle is an upper bound, not a competitor.}
  \label{tab:ladder_backbones}
  \small
  \setlength{\tabcolsep}{4pt}
  \begin{tabular}{lccccccccc}
    \toprule
    \multirow{2}{*}{Method} & \multicolumn{3}{c}{CE metrics}
    & \multicolumn{3}{c}{NLG metrics} & \multicolumn{3}{c}{Closed-ended VQA} \\
    \cmidrule(lr){2-4}\cmidrule(lr){5-7}\cmidrule(lr){8-10}
    & PREC & REC & F-1 & BLEU-4 & METEOR & ROUGE-L & Acc & F-1 & AUROC \\
    \midrule
    \multicolumn{10}{l}{\emph{LLaVA-1.5-7B}}\\[1pt]
    Zero-shot                  & 0.507 & 0.086 & 0.147 & 2.00 & 13.04 & 14.74 & 0.344 & 0.505 & 0.475 \\
    SFT                        & 0.656 & 0.629 & 0.642 & 25.21 & 42.81 & 50.92 & 0.631 & 0.487 & 0.665 \\
    $+$KR                      & 0.654 & 0.627 & 0.640 & 24.23 & 42.07 & 50.34 & 0.655 & \underline{0.542} & \underline{0.687} \\
    $+$CR                      & 0.720 & 0.746 & 0.733 & 29.50 & 47.32 & 52.61 & \textbf{0.693} & 0.328 & 0.649 \\
    \textbf{$+$SF (PerFact)}   & \underline{0.742} & \underline{0.785} & \textbf{0.763} & \underline{29.78} & \underline{47.53} & \textbf{53.97} & 0.648 & \textbf{0.573} & \textbf{0.689} \\
    $+$SF\,$+$KR               & 0.739 & \textbf{0.786} & \underline{0.762} & 29.61 & 47.14 & 52.94 & \underline{0.688} & 0.323 & 0.661 \\
    $+$SF\,$+$CR               & \textbf{0.756} & 0.752 & 0.754 & \textbf{30.38} & \textbf{48.02} & \underline{53.25} & 0.655 & 0.000 & 0.637 \\
    \emph{$+$SF$^*$ (oracle)}  & \emph{0.933} & \emph{0.831} & \emph{0.879} & \emph{37.11} & \emph{54.94} & \emph{59.08} & \emph{0.662} & \emph{0.040} & \emph{0.667} \\
    \midrule
    \multicolumn{10}{l}{\emph{InternVL3-8B}}\\[1pt]
    Zero-shot                  & 0.499 & 0.455 & 0.476 & 2.05 & 20.57 & 17.06 & 0.522 & 0.467 & 0.576 \\
    SFT                        & 0.717 & 0.690 & 0.703 & 27.33 & 45.24 & 52.37 & 0.635 & 0.588 & 0.716 \\
    $+$KR                      & 0.706 & 0.725 & 0.716 & 26.86 & 45.31 & 51.81 & 0.657 & \textbf{0.600} & \underline{0.723} \\
    $+$CR                      & 0.741 & 0.727 & 0.734 & 29.23 & 47.14 & 52.89 & 0.675 & 0.466 & 0.680 \\
    \textbf{$+$SF (PerFact)}   & \underline{0.754} & \textbf{0.808} & \textbf{0.780} & 28.98 & 47.18 & 53.07 & 0.663 & \underline{0.592} & 0.710 \\
    $+$SF\,$+$KR               & 0.746 & \underline{0.799} & 0.772 & \underline{29.32} & \textbf{47.70} & \underline{53.13} & \textbf{0.715} & 0.515 & \textbf{0.729} \\
    $+$SF\,$+$CR               & \textbf{0.778} & 0.771 & \underline{0.774} & \textbf{29.63} & \underline{47.50} & \textbf{53.37} & \underline{0.697} & 0.464 & 0.694 \\
    \emph{$+$SF$^*$ (oracle)}  & \emph{0.913} & \emph{0.884} & \emph{0.898} & \emph{36.25} & \emph{54.30} & \emph{58.39} & \emph{0.653} & \emph{0.571} & \emph{0.700} \\
    \bottomrule
  \end{tabular}
\end{table*}

The ordering is the paper's on all three: facts beat retrieval, adding retrieval to facts
changes little, and the oracle sits clear of every deployable row on report quality. What
differs is the starting point. InternVL3-8B fine-tunes to 0.703 ungrounded against 0.644 and
0.642, so the same fact sentence buys it $+0.076$ where it buys them $+0.131$ and $+0.121$;
the three still arrive within 0.017 of each other. The backbone sets how far the model must
travel, not where it ends up.

Question answering behaves as in \mainTabLadder{} on all three backbones: the grounding
source moves it far less than it moves report quality, and the oracle is not its best row --
ground-truth facts are worth $+0.104$ in clinical entity F1 on Qwen and $-0.054$ in area
under the curve. What the fact sentence does for the text, it does not do for the answer.
%

\subsection{The other three models}

\begin{figure}[!t]
  \centering
  \includegraphics[width=\columnwidth]{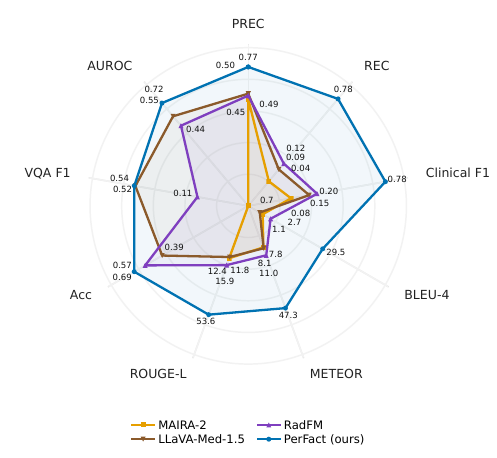}
  \caption{\textbf{The other three released models collapse the same way.} PerFact against
  MAIRA-2, LLaVA-Med-1.5 and RadFM, zero-shot on brain MRI, on the rings \mainFigRadar{}
  uses for the other three.}
  \label{fig:radar_rest}
\end{figure}

\mainFigRadar{} carries HuatuoGPT-Vision, LLaVA-Rad and CheXagent; Figure~\ref{fig:radar_rest}
adds the other three of the six.

With \mainFigRadar{}, the six divide by pretraining domain, not recency or size, and 3D
input alone does not transfer.

\section{Per-Disease Results}

\mainTabLadder{} aggregates the whole test set; Table~\ref{tab:perdisease_report} splits the
same runs by cohort, so no disease-specific result hides inside the aggregate. The All block
reproduces it exactly.

The two tumor cohorts are separated and the two non-tumor cohorts pooled, following the fact
schema (Section~D). The white-matter cohort contributes nine of the 47 pooled cases and does
not support a column of its own: at that size every method returns the same three values.


\begin{table*}[!t]
  \centering
  \caption{\textbf{Report generation by cohort.} Clinical entity scores for the eight conditions of \mainTabLadder{}, split by cohort. The two tumor cohorts are separated and acute stroke and white-matter hyperintensity (WMH) pooled, following the fact schema (Section~D). \textbf{Best} and \underline{second best} among deployable methods; the oracle row is an upper bound, not a competitor.}
  \label{tab:perdisease_report}
  \small
  \setlength{\tabcolsep}{4pt}
  \begin{tabular}{lcccccccccccc}
    \toprule
    \multirow{2}{*}{Method} & \multicolumn{3}{c}{Glioma ($n{=}40$)} & \multicolumn{3}{c}{Meningioma ($n{=}43$)} & \multicolumn{3}{c}{Stroke and WMH ($n{=}47$)} & \multicolumn{3}{c}{All ($n{=}130$)} \\
    \cmidrule(lr){2-4}\cmidrule(lr){5-7}\cmidrule(lr){8-10}\cmidrule(lr){11-13}
    & PREC & REC & F-1 & PREC & REC & F-1 & PREC & REC & F-1 & PREC & REC & F-1 \\
    \midrule
    Zero-shot                  & 0.771 & 0.638 & 0.698 & 0.427 & 0.506 & 0.463 & 0.325 & 0.214 & 0.258 & 0.541 & 0.480 & 0.509 \\
    SFT                        & 0.679 & 0.554 & 0.610 & 0.659 & 0.690 & 0.674 & 0.676 & 0.639 & 0.657 & 0.671 & 0.619 & 0.644 \\
    $+$KR                      & 0.662 & 0.501 & 0.571 & 0.707 & 0.729 & 0.718 & 0.676 & 0.613 & 0.643 & 0.682 & 0.601 & 0.639 \\
    $+$CR                      & 0.736 & 0.799 & 0.766 & 0.655 & \textbf{0.737} & 0.694 & 0.697 & 0.639 & 0.667 & 0.701 & 0.736 & 0.718 \\
    \textbf{$+$SF (PerFact)}   & \underline{0.790} & \textbf{0.858} & \textbf{0.822} & \textbf{0.720} & 0.725 & 0.723 & \textbf{0.799} & \underline{0.718} & \underline{0.757} & 0.771 & 0.779 & 0.775 \\
    $+$SF\,$+$KR               & 0.780 & \underline{0.858} & \underline{0.817} & \underline{0.712} & \underline{0.737} & \textbf{0.724} & \underline{0.795} & \textbf{0.748} & \textbf{0.771} & 0.763 & 0.791 & 0.777 \\
    $+$SF\,$+$CR               & \textbf{0.791} & 0.825 & 0.808 & 0.712 & 0.737 & \underline{0.724} & 0.790 & 0.664 & 0.721 & 0.766 & 0.754 & 0.760 \\
    \midrule
    \emph{$+$SF$^*$ (oracle)}  & \emph{0.964} & \emph{0.900} & \emph{0.931} & \emph{0.937} & \emph{0.871} & \emph{0.902} & \emph{0.805} & \emph{0.748} & \emph{0.776} & \emph{0.912} & \emph{0.849} & \emph{0.879} \\
    \bottomrule
  \end{tabular}
\end{table*}

\paragraph{The gain is largest where the facts are richest.}
On glioma, PerFact reaches 0.822 against 0.610 for supervised fine-tuning, a gain of 0.212.
There the fact sentence carries the most: a laterality, one or more lobes, three volumes, an
enhancement pattern and a category. On the pooled non-tumor cohorts, where it carries a lesion
and a burden band, the comparison is 0.757 against 0.657, a gain of 0.100.

\paragraph{The residual oracle gap tracks fact granularity, not cohort size.}
The gap between predicted and ground-truth facts is 0.108 on glioma and 0.180 on
meningioma, against 0.019 on the 47 pooled non-tumor cases. The non-tumor group is the largest
and has the smallest gap, so size does not drive it; what the schema asks upstream to get
right does. Three nested volumes and an enhancement pattern leave room for upstream error that
a single burden band does not. This is the per-cohort form of the main paper's aggregate
$-0.104$.

\paragraph{Meningioma is harder for every method.}
Every deployable condition scores lower on meningioma than glioma. The oracle reaches 0.902 there: with ground-truth facts
injected, meningioma is nearly as reportable as glioma. These runs do not isolate which
upstream field carries that cost.

The aggregate hides one boundary condition. On glioma, adding retrieval to the fact sentence
changes nothing, 0.817 against 0.822, inside the noise band; on the pooled non-tumor cohorts
it adds 0.014, just outside it. Coarse facts leave room that retrieval can fill; fine-grained
facts do not.

\section{Where the Gain Is Realized}

The grounding ladder moves clinical entity F1 score from 0.644 to 0.775, and the lift is not
spread evenly. Table~\ref{tab:entity} recomputes the per-entity breakdown from the same
stored predictions as \mainTabLadder, comparing supervised fine-tuning (SFT) against PerFact on
the same 130 cases with the same extractor.

\begin{table}[t]
\centering
\small
\setlength{\tabcolsep}{3.5pt}
\begin{tabular}{lrrrrr}
\toprule
Entity & $n$ & SFT & PerFact & Gain & Oracle \\
\midrule
\multicolumn{6}{l}{\emph{Disease category -- named by the fact sentence}}\\
glioma                & 40 & 0.204 & 0.902 & $+0.698$ & 0.841 \\
glioblastoma          & 39 & 0.208 & 0.889 & $+0.681$ & 0.800 \\
meningioma            & 41 & 0.643 & 0.902 & $+0.259$ & 0.965 \\
\midrule
\multicolumn{6}{l}{\emph{Location and laterality -- named by the fact sentence}}\\
temporal              & 31 & 0.323 & 0.698 & $+0.376$ & 0.737 \\
bilateral             & 22 & 0.063 & 0.296 & $+0.234$ & 0.500 \\
occipital             & 26 & 0.356 & 0.522 & $+0.166$ & 0.654 \\
left                  & 81 & 0.596 & 0.757 & $+0.161$ & 0.872 \\
right                 & 65 & 0.567 & 0.726 & $+0.159$ & 0.894 \\
parietal              & 40 & 0.364 & 0.519 & $+0.155$ & 0.757 \\
frontal               & 64 & 0.618 & 0.682 & $+0.064$ & 0.875 \\
cerebellum            & 20 & 0.410 & 0.323 & $-0.088$ & 0.545 \\
brainstem             & 12 & 0.174 & 0.353 & $+0.179$ & 0.667 \\
\midrule
\multicolumn{6}{l}{\emph{Mass effect and edema}}\\
midline shift         & 25 & 0.400 & 0.623 & $+0.223$ & 0.917 \\
ventricle compression & 30 & 0.500 & 0.667 & $+0.167$ & 0.918 \\
edema                 & 53 & 0.800 & 0.920 & $+0.120$ & 0.925 \\
extensive edema       &  6 & 0.400 & 0.222 & $-0.178$ & 1.000 \\
\midrule
\multicolumn{6}{l}{\emph{Already at ceiling under fine-tuning alone}}\\
midline               &129 & 0.996 & 0.992 & $-0.004$ & 1.000 \\
enhancing             & 80 & 0.982 & 0.982 & $\pm0.000$ & 0.982 \\
infarct               & 36 & 0.946 & 0.930 & $-0.016$ & 0.959 \\
\midrule
ring enhancing        &  9 & 0.235 & 0.333 & $+0.098$ & 0.947 \\
non-enhancing         &  3 & 0.000 & 0.000 & $\pm0.000$ & 0.000 \\
\bottomrule
\end{tabular}
\caption{\textbf{The gain lands on the fields the fact sentence carries.} Clinical entity
F1 by entity over 130 cases; $n$ is how many references name it (Section~F). Fifteen improve
by more than 0.02, two worsen, four move less. Oracle is the same measurement with
ground-truth facts.}
\label{tab:entity}
\end{table}

\paragraph{The gain concentrates where the fact sentence supplies content.}
Disease category moves furthest: glioma and glioblastoma each rise by roughly $+0.69$ from a
base near 0.20. Fine-tuning on images alone barely distinguishes the two; the fact sentence
states the predicted category outright. Laterality and lobe, the other fields the serializer
always emits, move by $+0.06$ to $+0.38$, cerebellum excepted. Mass-effect entities gain as
much without being named in the fact sentence.

\paragraph{Entities at ceiling do not move.}
The three entities already above 0.94 under fine-tuning alone move by at most 0.016, and
downward: they are in almost every report (midline, 129 of 130) or inferable from the
images.

\paragraph{Two entities get worse, both on thin support.}
Cerebellum falls by 0.088 on 20 cases and extensive edema by 0.178 on six. Coarse fields
mislead: a lesion whose centroid sits near the midline can be serialized away from an
infratentorial description, and the burden band that produces ``extensive'' is a threshold
on a predicted volume. Neither regression is large against its support, but they are the two
places where injected grounding hurts.

The oracle column locates the remaining headroom, largest where the schema asks upstream for
the most detail: ring enhancement 0.333 against 0.947, midline shift 0.623 against 0.917. Section~J locates the same gap by cohort.

\section{Three Cases End to End}

\mainFigHallucination{} prints one held-out study and what each system wrote about it.
Table~\ref{tab:qualitative} does the same for three glioma cases, chosen by rank rather
than by eye: the references asserting the most, the median and the fewest of the 24
clinical entities. That ranking is a property of the study and of no system's output, so
it cannot favor or embarrass a particular row.

\definecolor{qualok}{RGB}{0,120,60}
\definecolor{qualbad}{RGB}{190,30,30}
\begin{table*}[!t]
\centering
\small
\setlength{\tabcolsep}{4pt}
\begin{tabular}{@{}l r p{0.68\textwidth}@{}}
\toprule
System & F1 (ents.) & Generated report, verbatim \\
\multicolumn{3}{@{}l@{}}{\emph{Hardest case}, reference asserts 13 of the 24 entities: ``Findings: A lesion with approximately round abnormal signal is seen in\,\dots''} \\[2pt]
\midrule
HuatuoGPT-Vision & $0.636$ (9) & \texttt{\footnotesize hyperintense lesion in the \textcolor{qualok}{left} \textcolor{qualbad}{parietal} lobe on T2-weighted\,\dots} \\
LLaVA-Rad & $0.267$ (2) & \texttt{\footnotesize provided. Multiple clips in the \textcolor{qualok}{left} axilla with a \textcolor{qualok}{left} breast\,\dots} \\
Qwen2.5-VL zero-shot & $0.636$ (9) & \texttt{\footnotesize heterogeneous mass in the \textcolor{qualok}{right} \textcolor{qualbad}{parietal} lobe. - The mass\,\dots} \\
SFT & $0.917$ (11) & \texttt{\footnotesize signal focus is visible in the \textcolor{qualok}{left} \textcolor{qualok}{frontal} lobe, presenting\,\dots} \\
$+$CR & $0.870$ (10) & \texttt{\footnotesize signal focus is observed in the \textcolor{qualok}{left} \textcolor{qualok}{frontal} lobe, displaying a\,\dots} \\
\textbf{$+$SF (PerFact)} & $0.960$ (12) & \texttt{\footnotesize signal focus is observed in the \textcolor{qualok}{left} \textcolor{qualok}{frontal} lobe, presenting\,\dots} \\
$+$SF$^*$ oracle & $0.917$ (11) & \texttt{\footnotesize signal focus is visible in the \textcolor{qualok}{left} \textcolor{qualok}{frontal} lobe, presenting\,\dots} \\
\midrule
\multicolumn{3}{@{}l@{}}{\emph{Median case}, reference asserts 9 of the 24 entities: ``Findings: Multiple clustered abnormal signal shadows are seen in the\,\dots''} \\[2pt]
\midrule
HuatuoGPT-Vision & $0.800$ (6) & \texttt{\footnotesize well-circumscribed lesion in the \textcolor{qualok}{right} cerebral hemisphere. -\,\dots} \\
LLaVA-Rad & $0.200$ (1) & \texttt{\footnotesize intact. No free air below the \textcolor{qualok}{right}\,\dots} \\
Qwen2.5-VL zero-shot & $0.533$ (6) & \texttt{\footnotesize heterogeneous mass in the \textcolor{qualbad}{left} \textcolor{qualbad}{parietal} lobe. - The mass\,\dots} \\
SFT & $0.375$ (7) & \texttt{\footnotesize signal focus is observed in the \textcolor{qualbad}{left} \textcolor{qualbad}{cerebellopontine} angle\,\dots} \\
$+$CR & $0.900$ (11) & \texttt{\footnotesize signal focus is observed in the \textcolor{qualok}{right} \textcolor{qualok}{frontal} lobe, presenting\,\dots} \\
\textbf{$+$SF (PerFact)} & $0.900$ (11) & \texttt{\footnotesize signal focus is observed in the \textcolor{qualok}{right} \textcolor{qualok}{frontal} lobe, presenting\,\dots} \\
$+$SF$^*$ oracle & $0.824$ (8) & \texttt{\footnotesize signal focus is observed in the \textcolor{qualok}{right} \textcolor{qualok}{frontal} lobe, presenting\,\dots} \\
\midrule
\multicolumn{3}{@{}l@{}}{\emph{Easiest case}, reference asserts 6 of the 24 entities: ``Findings: A mass-like abnormal signal is observed within the posterior\,\dots''} \\[2pt]
\midrule
HuatuoGPT-Vision & $0.706$ (11) & \texttt{\footnotesize hyperintense lesion located in the \textcolor{qualok}{left} \textcolor{qualbad}{temporal} lobe. The\,\dots} \\
LLaVA-Rad & $0.250$ (2) & \texttt{\footnotesize significant collapse of the \textcolor{qualbad}{right} lung. \textcolor{qualok}{Left} lung is clear.\,\dots} \\
Qwen2.5-VL zero-shot & $0.833$ (6) & \texttt{\footnotesize heterogeneous mass in the \textcolor{qualok}{left} \textcolor{qualbad}{parietal} lobe. - The mass\,\dots} \\
SFT & $0.500$ (6) & \texttt{\footnotesize signal is observed in the \textcolor{qualbad}{right} \textcolor{qualbad}{temporal} lobe, exhibiting an\,\dots} \\
$+$CR & $0.750$ (10) & \texttt{\footnotesize signal focus is observed in the \textcolor{qualok}{left} \textcolor{qualbad}{temporal} lobe, presenting\,\dots} \\
\textbf{$+$SF (PerFact)} & $0.800$ (9) & \texttt{\footnotesize signal focus is visible in the \textcolor{qualok}{left} \textcolor{qualbad}{temporal} lobe, presenting\,\dots} \\
$+$SF$^*$ oracle & $1.000$ (6) & \texttt{\footnotesize signal focus can be seen in the \textcolor{qualok}{left} lateral ventricle,\,\dots} \\
\bottomrule
\end{tabular}
\caption{\textbf{Three glioma cases, ranked by how much there is to report.} The reference asserting the most, the median and the fewest of the 24 entities, over the 40 glioma cases; F1 is per case, the count in parentheses how many entities that report asserts. Color marks each word the extractor keys on, green where the reference asserts it too. The green \emph{left} in LLaVA-Rad's chest report is not an error: the extractor reads laterality without regard to what it qualifies, the inflation the paper notes in quoting that row's recall. Over the 40 cases $+$SF beats $+$CR on 24, ties 8, loses 8. Excerpts are verbatim prefixes; \dots\ marks an elision.}
\label{tab:qualitative}
\end{table*}

Two failures are categorical rather than graded, which no column of means shows. LLaVA-Rad
answers a brain study with a chest report on all three, and the zero-shot backbone opens all
three with the same sentence: part of what it writes is a template the image never reaches.

\section{Reproducibility Notes}

\subsection{Computing environment}

Every reported run used one NVIDIA GH200 (96\,GB), 72 CPU cores and 115\,GB host memory on
one Slurm node, with no sharding, DeepSpeed or FSDP. The host is an ARM Neoverse-V2 running
SUSE Linux Enterprise Server 15 SP6, kernel 6.4.0, aarch64. The GPU driver was not logged.

Generator and released baselines: Python 3.11, CUDA 12.6, PyTorch 2.12.1, torchvision
0.27.1, Transformers 5.6.0, PEFT 0.18.1, Accelerate 1.11.0, TRL 0.24.0, Datasets 4.0.0,
tokenizers 0.22.2, NumPy 2.4.4, and LLaMA-Factory 0.9.6.dev0 at commit \texttt{666ee0ca}.
Upstream perception runs separately on the same PyTorch build plus MONAI 1.5.2, nibabel
5.4.2, open\_clip 3.3.0, rank-bm25 0.2.2, scikit-learn 1.9.0 and SciPy 1.17.1. Baselines pin
their own Transformers versions: 4.36.2 for LLaVA-Med, 4.31.0 for LLaVA-Rad, 4.40.0.dev0 for
HuatuoGPT-Vision, 4.51.3 for MAIRA-2 and 4.28.1 for RadFM.

Wall-clock on one GPU: the three upstream curriculum stages 1\,h\,21\,m, 9\,h\,59\,m and
9\,h\,49\,m; a generator fine-tune about six minutes; the 130-case test set five to ten
minutes; external validation over 298 studies under four hours. A LoRA adapter is 639\,MB,
the upstream multi-task network 34.8\,MB, each binary segmenter 18.8\,MB.

\subsection{Determinism}

Seed 42 fixes the data split, the attribute probe's cross-validation folds, the VQA mixture
sampling and all upstream training, where PyTorch, NumPy, Python and MONAI's determinism
switch are set. The generator is reseeded at 42, 1234 and 2024 for the backbone comparison;
everything else uses 42.

Decoding is greedy for every reported number, released baselines included (Section~H).

\subsection{Which file backs which number}

Where a quantity was measured more than once, the paper fixes one file: the grounding ladder
from the axis-A evaluation, the per-disease breakdown from the full evaluation of the same
predictions, the backbone comparison from the three-seed summary, and the matching recall
from the second matching run rather than the first.

\subsection{Code archive}

The archive holds the code behind every reported number, the split assignment for every
case, 95 stored metric files, the per-case reports for all fourteen conditions and the
per-question scores for the closed-ended bank. Its README lists the contents.

\texttt{scripts/check\_tables.py} needs only the standard library, prints paper against
stored value for the grounding ladder, the backbone comparison, the upstream segmentation
results and the two breakdowns of Section~E, and recomputes the ANOVA from the fifteen
per-seed scores rather than quoting it; on the shipped archive it reports 36 of 36 in
agreement.

No imaging is redistributed (Section~B); the white-matter prohibition on derived data keeps
cached slices and embeddings out, and checkpoints are omitted for size. What ships
recomputes every reported number without a GPU and rebuilds the pipeline from the original
sources. The code is MIT-licensed; the datasets are not.

\bibliography{references}